\documentclass[final,5p,times,twocolumn,authoryear]{elsarticle}
\let\oldparagraph\paragraph
\renewcommand{\paragraph}[1]{\oldparagraph{\textbf{#1}}}

\usepackage{amssymb}
\usepackage{booktabs}
\usepackage{multirow}
\usepackage{graphicx}
\usepackage{dblfloatfix}
\usepackage{amsmath}
\usepackage{xurl}
\usepackage[T1]{fontenc}
\usepackage{hyperref}
\usepackage{orcidlink}
\hypersetup{
    colorlinks=true,
    linkcolor=blue,
    citecolor=blue,
    urlcolor=blue
}

\journal{ISPRS Journal of Photogrammetry and Remote Sensing}

\begin{document}

\begin{frontmatter}



\title{STARS-GS: Structure-Aware Regularized Gaussian Splatting for Large-Scale Aerial Surface Reconstruction} 


\author[inst2,inst3]{Bocheng~Li\,\orcidlink{0009-0000-7543-2914}}

\author[inst1]{Wenjuan~Zhang\corref{cor1}\,\orcidlink{0000-0002-0534-0974}}
\ead{zhangwj@aircas.ac.cn}
\cortext[cor1]{Corresponding author}

\author[inst1]{Jie~Pan\,\orcidlink{0009-0001-2548-4684}}
\author[inst2]{Dongxu~Han\,\orcidlink{0009-0007-7102-8050}}
\author[inst2,inst3]{Xuesong~Ma\,\orcidlink{0009-0002-2329-3750}}
\author[inst2,inst3]{Yiling~Yao\,\orcidlink{0009-0009-1025-8674}}
\author[inst2,inst3]{Yaning~Wang\,\orcidlink{0009-0000-1801-904X}}

\affiliation[inst1]{organization={State Key Laboratory of Remote Sensing and Digital Earth, Aerospace Information Research Institute, Chinese Academy of Sciences},
            addressline={},
            city={Beijing},
            postcode={100101},
            state={},
            country={China}}

\affiliation[inst2]{organization={Aerospace Information Research Institute, Chinese Academy of Sciences},
            addressline={},
            city={Beijing},
            postcode={100094},
            state={},
            country={China}}

\affiliation[inst3]{organization={University of Chinese Academy of Sciences},
            addressline={},
            city={Beijing},
            postcode={100049},
            state={},
            country={China}}

\begin{abstract}
Large-scale 3D surface reconstruction from aerial imagery is fundamental to geospatial mapping and urban modeling. Recent advances in 3D Gaussian Splatting (3DGS) have demonstrated considerable potential for this task. However, existing methods still face three major challenges in large and complex scenes: scene partitioning may split continuous scene elements across independently optimized sub-regions; geometric constraints mainly focus on the attributes of individual Gaussians while overlooking their local organization; and uniform regularization struggles to accommodate heterogeneous geometric structures. To address these issues, we propose STARS-GS, a structure-aware 3DGS framework for large-scale surface reconstruction. First, we introduce a structure-aware scene partitioning strategy that better preserves continuous scene structures during partitioning and reduces cross-region geometric inconsistencies and stitching artifacts through boundary refinement. Second, we develop neighborhood-aware Gaussian organization that extends geometric constraints from individual primitives to their neighborhood organization, encouraging Gaussians to better conform to local surface geometry. Third, we introduce adaptive surface regularization that adjusts the regularization strength according to local geometric characteristics, promoting geometric consistency in structured regions while preserving plausible variations in unstructured regions. Extensive experiments on large-scale aerial photogrammetry benchmarks demonstrate that STARS-GS consistently outperforms the evaluated Gaussian-based methods in surface reconstruction. It  increases the average F1-score from 0.640 for the second-best method to 0.698, corresponding to a relative improvement of approximately 9.1\%, demonstrating effective improvements in geometric accuracy and surface completeness.
\end{abstract}


\end{frontmatter}



\section{Introduction}
\label{Introduction}


Large-scale surface reconstruction is a fundamental and challenging task in photogrammetry and Earth observation. It aims to recover geometrically accurate and structurally continuous 3D representations of terrain and above-ground objects from multi-view aerial images \citep{christodoulides2025survey,wu2024multiview3d,huang2024surfacereconstruction}, with critical applications in digital twins, urban planning, terrain analysis and other geospatial applications \citep{biljecki2015applications,pan2024heritagedigitaltwin}. Building upon the camera poses and sparse point clouds generated through Structure-from-Motion (SfM) \citep{schonberger2016sfm}, traditional pipelines based primarily on Multi-View Stereo (MVS) \citep{schonberger2016mvs} further derive dense point clouds through correspondence matching and subsequently extract surface meshes using dedicated surface reconstruction algorithms \citep{kazhdan2006poisson,kazhdan2013screenedpoisson}. Despite their proven effectiveness, traditional pipelines may still suffer from incomplete geometry and local structural distortions due to unreliable correspondences \citep{zhu2015occlusionmvs,liao2024highcompleteness}. Moreover, dense matching over large collections of high-resolution images incurs substantial computational overhead, limiting overall efficiency and scalability \citep{furukawa2010internetmvs}.

As an emerging approach to 3D reconstruction, 3D Gaussian Splatting (3DGS) explicitly represents scenes using learnable Gaussian primitives and employs an efficient differentiable rasterization pipeline, thereby enabling photorealistic rendering in real time \citep{kerbl2023gaussians}. Recent studies have further extended 3DGS to large-scale scenes through partitioned and parallel training strategies \citep{lin2024vastgaussian,wu2025blockgaussian,liu2024citygaussian}. In parallel, surface-oriented variants have demonstrated the capability of extracting surface meshes from Gaussian representations \citep{guedon2024sugar,huang2024twodgs,chen2025pgsr}. These developments indicate the potential of 3DGS for large-scale 3D surface reconstruction \citep{liu2025citygaussianv2,li2025ulsrgs,xiang2026gaussiancraft}.

However, existing GS-based methods still face several challenges when applied to complex, large-scale surface reconstruction \citep{chen2026survey3dgs}. (1) As scene scale increases, maintaining reconstruction quality generally requires more Gaussian primitives, leading to increased computational and GPU memory demands \citep{christodoulides2025survey,bao2025survey3dgs}. Large-scale GS methods therefore commonly partition the scene into independently optimized sub-regions based on camera visibility or point-cloud density \citep{lin2024vastgaussian,liu2024citygaussian,wu2025blockgaussian}. However, these strategies are primarily designed to control the amount of data processed during optimization, without adequately accounting for the structural integrity of scene elements \citep{wang2025pgsag,han2026vopgs}. Consequently, continuous structures may be divided across sub-regions with different observations and optimized independently without unified cross-region geometric constraints, leading to structural distortions and stitching artifacts near partition boundaries \citep{zhu2026sceneaware,wang2026consistencypreserving}. (2) Vanilla 3DGS primarily optimizes Gaussian primitives using rendering losses and lacks sufficient explicit geometric constraints, making it difficult to ensure the geometric fidelity of reconstructed surfaces \citep{turkulainen2025dnsplatter,gao2025surfacesplat}. To alleviate this limitation, existing methods incorporate depth and normal constraints, however, these constraints primarily focus on the attributes of individual primitives \citep{huang2024twodgs,dai2024gaussiansurfels,chen2025pgsr}. A continuous surface, however, is determined collectively by multiple neighboring Gaussian primitives, and its geometry depends not only on individual primitive attributes but also on their relative spatial arrangement \citep{huang2024surfacereconstruction,li2024geogaussian,shen2025topologygs,pan2026topogs}. In large-scale scenes, complex spatial layouts make such relationships more difficult to capture \citep{xiang2026gaussiancraft,li2025ulsrgs}. Therefore, optimizing individual primitive attributes is insufficient to recover local surface structures consistent with the actual geometry \citep{pan2026topogs,jiang2026halogs}. (3) Large-scale scenes contain heterogeneous scene elements, including buildings, roads, vegetation, and bare ground, which differ substantially in geometric and structural complexity \citep{liu2025citygaussianv2,yao2026arsgaussian}. However, most existing 3DGS methods apply a uniform surface regularization strategy across the entire scene \citep{huang2024twodgs,chen2025pgsr} and therefore struggle to simultaneously recover accurate surfaces in structured regions while preserving plausible local geometric variations in unstructured regions\citep{xiong2026adaptivecontrol,lian2026depthgradient,ververas2024sags},often causing structural distortions or boundary expansion in structured regions and oversmoothing in unstructured regions.

To address these challenges, we propose STARS-GS, a structure-aware 3DGS framework for large-scale surface reconstruction that improves geometric consistency across scene, neighborhood, and local-region scales. First, structure-aware scene partitioning uses spatial proximity and surface-normal similarity, supplemented by color, to group spatially connected and structurally similar areas within the same sub-region while aligning partition boundaries with scene-element boundaries, augmenting the partitioning objective to jointly control training scale and preserve scene-element integrity. Boundary refinement then aligns the geometry and appearance of adjacent sub-regions, reducing geometric misalignment and stitching artifacts caused by independent sub-region optimization and subsequent merging. Second, within each sub-region, neighborhood-aware Gaussian organization constructs a geometry-aware neighborhood graph based on spatial proximity and geometric similarity and dynamically updates the graph during 3DGS densification. Based on this graph, normal-consistency and tangential-distribution constraints jointly regulate the relative organization of neighboring primitives by promoting coherent normal orientations, guiding Gaussian centers toward local surface tangent planes and suppressing normal-direction stacking, respectively. Finally, adaptive surface regularization adjust the strengths of surface regularization according to local planarity, promoting geometric consistency in structured regions while preserving plausible geometric variations in unstructured regions. 

By improving the geometric accuracy, structural continuity, and scalability of aerial surface reconstruction, STARS-GS aims to provide a more reliable 3D scene representation for large-scale geospatial mapping. Our contributions are summarized as follows:

\begin{enumerate}
    \item We propose a structure-aware scene partitioning strategy that better preserves the structural integrity of scene elements across sub-regions and maintains cross-region surface continuity.

    \item We propose a neighborhood-aware Gaussian organization method that complements conventional primitive constraints with neighborhood spatial constraints, improving the geometric fidelity of reconstructed surfaces.

    \item We propose an adaptive surface regularization term that adjusts constraint strength according to local geometric characteristics, balancing geometric consistency and local variation across heterogeneous regions. 
\end{enumerate}

\section{Related Work}
\label{Related Work}

\subsection{Large-Scale 3D Reconstruction}
\label{subsec:large}

Large-scale 3D reconstruction involves extensive spatial coverage and large numbers of multi-view images, making GPU memory and computation major challenges \citep{christodoulides2025survey,tancik2022blocknerf,turki2022meganerf}. Early studies mainly focused on scaling Neural Radiance Fields (NeRF) \citep{mildenhall2020nerf} to large-scale scenes. Block-NeRF \citep{tancik2022blocknerf} and Mega-NeRF \citep{turki2022meganerf} partition large-scale scenes into sub-regions and train separate models. BungeeNeRF \citep{xiangli2022bungeenerf} adopts a progressive representation for multi-scale observations; Switch-NeRF \citep{mi2023switchnerf} employs a mixture-of-experts framework for scene decomposition; and Grid-NeRF \citep{xu2023gridnerf} combines NeRF with multi-resolution feature grids. Despite improved scalability, their implicit representations and volumetric rendering pipelines still incur substantial training and inference costs, limiting efficient large-scale reconstruction and real-time rendering \citep{kerbl2023gaussians}.

Benefiting from its differentiable rasterization pipeline, 3DGS \citep{kerbl2023gaussians} provides a more efficient alternative for large-scale reconstruction. Existing large-scale 3DGS methods typically partition the scene into sub-regions, assign corresponding training views, and optimize each sub-region independently, thereby reducing GPU memory and computation per  sub-region\citep{chen2026survey3dgs,lin2024vastgaussian,wu2025blockgaussian}.

VastGaussian \citep{lin2024vastgaussian} introduces a progressive partitioning strategy that distributes training cameras and point clouds across sub-regions according to an airspace-aware visibility criterion. BlockGaussian \citep{wu2025blockgaussian} adaptively determines block extents according to regional complexity and introduces auxiliary points to alleviate supervision mismatch during independent block training. CityGaussian \citep{liu2024citygaussian} uses a global scene prior and adaptive training data selection to support efficient block-wise training and fusion. These methods primarily control sub-region scale and computational load, making large-scale 3DGS training tractable \citep{chen2026survey3dgs,bao2025survey3dgs}.

Recent studies have increasingly focused on improving the geometric quality of large-scale surface reconstruction.. CityGaussianV2 \citep{liu2025citygaussianv2} extends CityGaussian with geometry-oriented 2D Gaussian modeling and optimization. ULSR-GS \citep{li2025ulsrgs} combines point-to-photo partitioning with multi-view-guided densification and cross-view geometric consistency constraints. GaussianCraft \citep{xiang2026gaussiancraft} integrates visibility-in-block selection with fine-grained geometric refinement.

However, even in recent geometry-oriented methods, geometric refinement is mainly performed within the resulting sub-regions, while scene partitioning is still primarily designed for computational efficiency and scalability, with limited consideration of how partition boundaries affect reconstruction geometry. Consequently, partition boundaries may cut through geometrically continuous structures, causing different parts of the same surface to be optimized separately, resulting in cross-region geometric inconsistencies and stitching artifacts \citep{liu2025mixgs,fan2025momentumgs,zhu2026sceneaware,wang2026consistencypreserving}.

\begin{figure*}[t]
    \centering
    \includegraphics[width=\linewidth]{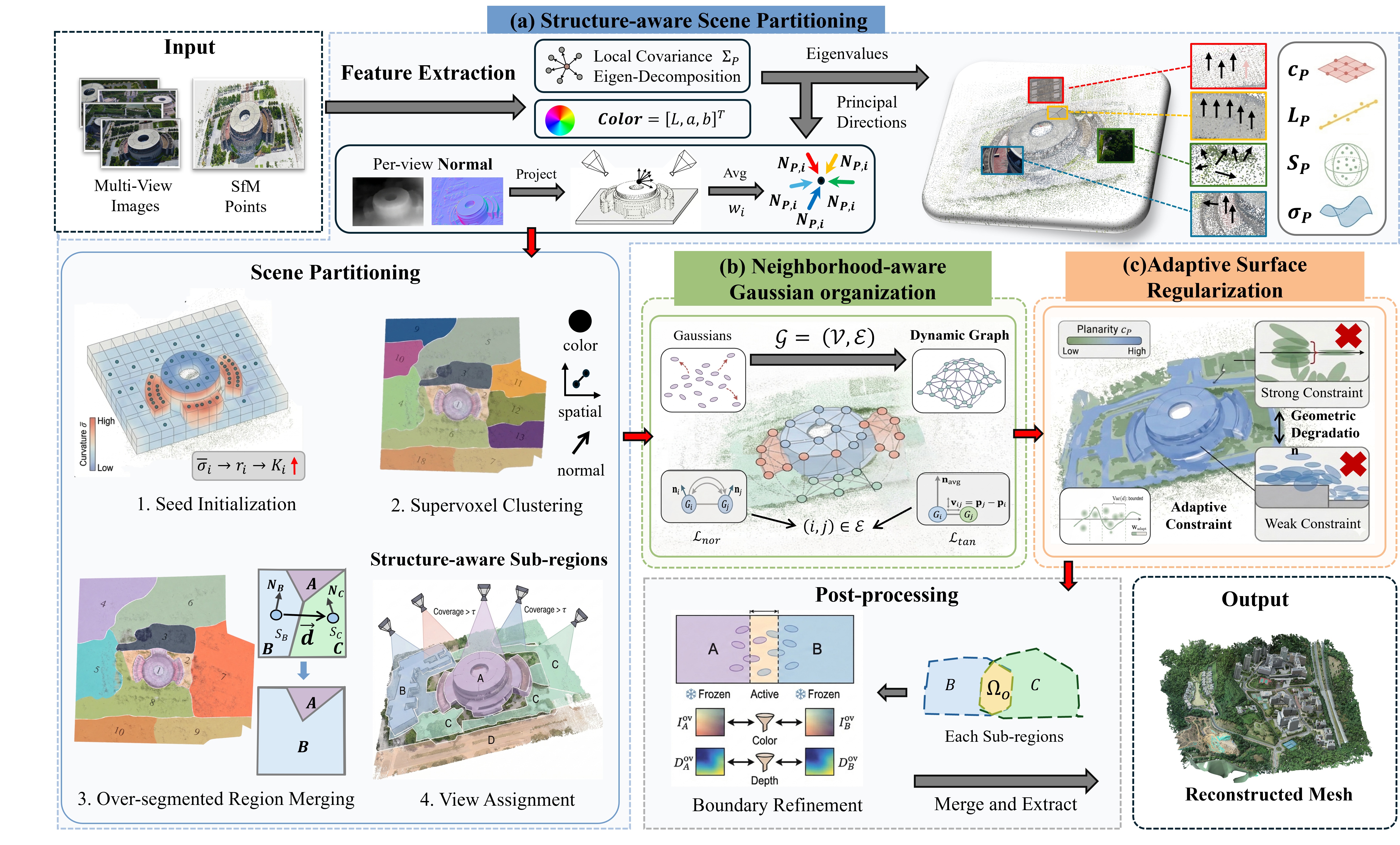}
    \caption{
    Overall workflow of STARS-GS. Given multi-view aerial images and SfM point cloud, the pipeline first constructs local geometric features, robust normals, and color information for \textbf{(a) structure-aware scene partitioning}, including curvature-guided seed initialization, supervoxel clustering, RAG-based region merging, and view assignment. Each sub-region is then optimized with \textbf{(b) neighborhood-aware Gaussian organization} and \textbf{(c) adaptive surface regularization}, which respectively maintain local Gaussian organization and adjust surface regularization according to local planarity. After independent sub-region optimization, boundary refinement is performed within overlap bands before sub-region merging and surface mesh extraction.
    }
    \label{fig:pipeline}
\end{figure*}

\subsection{Multi-View Surface Reconstruction}
\label{subsec:surface}

Conventional SfM-MVS reconstruction pipelines \citep{schonberger2016sfm,schonberger2016mvs} remain sensitive to unreliable correspondence matching, which can cause incomplete geometry and structural distortions in weakly textured regions, non-Lambertian surfaces, and severe occlusions \citep{aanaes2016mvsdata,yamashita2023nlmvs,zhu2015occlusionmvs}. Neural implicit methods, including NeuS \citep{wang2021neus} and Neuralangelo \citep{li2023neuralangelo},  improve the surface representation using signed distance fields (SDFs) or multi-resolution hash \citep{yariv2021volsdf,muller2022instantngp}. However, their volumetric rendering pipelines incur substantial computational overhead \citep{xie2022neuralfields}.

3DGS combines high-fidelity appearance reconstruction with efficient rendering through explicit Gaussian primitives and differentiable rasterization \citep{kerbl2023gaussians}. However, its discrete and unstructured primitives do not inherently conform to continuous physical surfaces \citep{huang2024twodgs,dai2024gaussiansurfels}. Consequently, subsequent studies have explored two main directions: indirect improvement through appearance modeling and direct geometric constraints.

Appearance-based methods reduce appearance-geometry ambiguities by better modeling view-dependent effects. Ref-GS \citep{zhang2025refgs}, Reflections Unlock (Ref-Unlock) \citep{song2025refunlock}, and RTR-GS \citep{zhou2025rtrgs} employ directional light factorization, reflection disentanglement, and BRDF-lighting decomposition, respectively. However, their primary focus on appearance provides only indirect improvements to surface geometry \citep{ge2026gipgs,li2026rags}.

By contrast, direct geometric enhancement methods introduce explicit geometric priors or constraints. GS2Mesh \citep{wolf2024gs2mesh} uses a pretrained stereo-matching network after 3DGS optimization to recover surface geometry. TrimGS \citep{fan2024trimgs} removes geometrically unreliable primitives according to their image contributions and employs scale-driven densification to control Gaussian size. SuGaR \citep{guedon2024sugar} introduces surface-alignment regularization to align Gaussians with the local scene surface. GSDF \citep{yu2024gsdf} introduces an SDF branch to regulate Gaussian density using SDF predictions and jointly optimizes the 3DGS and SDF branches through mutual guidance. 2DGS \citep{huang2024twodgs} replaces volumetric 3D Gaussians with oriented 2D Gaussian disks for surface-aligned representation. PGSR \citep{chen2025pgsr} derives unbiased depth from Gaussian-plane distance and normal maps to improve multi-view depth and normal consistency.

However, external networks and additional geometric constraints may introduce generalization or scalability concerns \citep{wolf2024gs2mesh,zhang2022stereogeneralization,jiang2026halogs}. Moreover, most existing constraints focus on the attributes of individual Gaussian primitives\citep{bao2025survey3dgs}, whereas the spatial organization among neighboring primitives remains insufficiently constrained. In addition, existing methods commonly apply uniform surface regularization, without accounting for the different surface characteristics of structured and unstructured regions \citep{guedon2024sugar,huang2024twodgs,chen2025pgsr}, a limitation that becomes more pronounced in large-scale scenes containing heterogeneous surface structures\citep{yao2026arsgaussian}.

\section{Method}
\label{Method}

As illustrated in Fig.~\ref{fig:pipeline}, given multi-view images and a sparse SfM point cloud, STARS-GS first divides the scene into structurally coherent sub-regions through structure-aware partitioning (Section~\ref{subsec:structure_partitioning}). Each sub-region is then optimized using neighborhood-aware Gaussian organization (Section~\ref{subsec:topology_organization}) and adaptive surface regularization (Section~\ref{subsec:adaptive_surface_regularization}) to improve local Gaussian organization and surface geometry. After independent optimization, shared boundary regions are refined to improve cross-region consistency before sub-region merging and surface extraction.

\subsection{Preliminaries}
\label{subsec:3dgs}

3DGS represents a scene using learnable 3D Gaussian primitives \citep{kerbl2023gaussians}. Each primitive is parameterized as $\mathcal{G}_i=(\boldsymbol{\mu}_i,\boldsymbol{\Sigma}_i,\mathbf{c}_i,\alpha_i)$, where $\boldsymbol{\mu}_i \in \mathbb{R}^{3}$ denotes its position, $\boldsymbol{\Sigma}_i \in \mathbb{R}^{3 \times 3}$ encodes its spatial extent and orientation, $\mathbf{c}_i$ its color attributes, and $\alpha_i$ its opacity.

After projecting and depth-sorting the Gaussians \citep{zwicker2001surface}, the color at pixel $\mathbf{x}$ is rendered by differentiable $\alpha$-compositing:

\begin{equation}
\begin{aligned}
    a_i(\mathbf{x})
    &= \alpha_i G_i(\mathbf{x}), \\
    T_i(\mathbf{x})
    &= \prod_{j=1}^{i-1}
       \left(1-a_j(\mathbf{x})\right), \\
    \mathbf{C}(\mathbf{x})
    &= \sum_{i=1}^{N}
       T_i(\mathbf{x})\,a_i(\mathbf{x})\,\mathbf{c}_i .
\end{aligned}
\label{eq:alpha_compositing}
\end{equation}

where $G_i(\mathbf{x})$ denotes the projected 2D Gaussian response at pixel $\mathbf{x}$, while $a_i(\mathbf{x})$ and $T_i(\mathbf{x})$ denote the effective opacity and accumulated transmittance, respectively. For convenience, we denote the $\alpha$-compositing contribution of Gaussian $i$ to pixel $\mathbf{x}$ as $\omega_i(\mathbf{x})=T_i(\mathbf{x})a_i(\mathbf{x})$.

Optimization minimizes the rendering loss between the rendered image $\hat{I}$ and ground truth $I$ \citep{kerbl2023gaussians,wang2004ssim}:

\begin{equation}
    \mathcal{L}_{\mathrm{render}}
    =
    \left(1-\lambda_{\mathrm{ssim}}\right)
    \left\|\hat{I}-I\right\|_{1}
    +
    \lambda_{\mathrm{ssim}}
    \left[
    1-\operatorname{SSIM}\left(\hat{I},I\right)
    \right].
    \label{eq:rendering_loss}
\end{equation}

However, rendering loss alone does not adequately constrain surface geometry and may therefore yield geometrically inconsistent surfaces \citep{huang2024twodgs,chen2025pgsr,jiang2026halogs}.

\subsection{Structure-Aware Scene Partitioning}
\label{subsec:structure_partitioning}

Conventional spatial partitioning may split continuous scene elements across sub-regions, causing different parts of them to be optimized independently and leading to introducing geometric inconsistencies and stitching artifacts. To address this issue, we propose a structure-aware scene partitioning method. Specifically, each valid SfM point is characterized by local geometric features, a robust normal, and color features. Based on this representation, structure-aware supervoxel clustering is then performed \citep{papon2013vccs}, encouraging spatially connected points with similar structural characteristics to be grouped into locally coherent regions. Structurally similar adjacent regions are further merged using a RAG to reduce residual over-segmentation, followed by training-view assignment for independent sub-region optimization. After all sub-regions have been optimized independently, boundary consistency refinement is performed over the overlapping areas between adjacent sub-regions.

Sparse SfM points often contain unreliable points caused by correspondence mismatches and occlusions. Directly including these points in local feature computation may lead to unreliable results. Before feature construction,  we retain only points satisfying $\eta_P \geq \tau_{\mathrm{track}}$, where $\eta_P$ is the number of valid views in the SfM track, yielding $\mathcal{P}{\mathrm{valid}}={P\in\mathcal{P}{\mathrm{raw}}\mid\eta_P\ge\tau_{\mathrm{track}}}$.

\subsubsection{Joint Feature Construction}
\label{subsubsec:joint_features}

As illustrated in Fig.~\ref{fig:pipeline}(a), for each valid SfM point $P\in\mathcal{P}_{\mathrm{valid}}$, we construct a joint feature comprising local geometric features, robust normal, and color.

\paragraph{Local geometric feature construction} We first query each valid point spatial $k$-nearest-neighbor set $\mathcal{N}_k(P)$ within a maximum search radius and compute the local covariance matrix \citep{hackel2016contour}. Eigenvalue decomposition yields $\lambda_1\geq\lambda_2\geq\lambda_3$, and the eigenvector $\mathbf{e}_3$ of the smallest eigenvalue is used as the initial normal \citep{hoppe1992surface}. Based on them, we derive planarity $c_P$, curvature $\sigma_P$, linearity $\ell_P$, and sphericity $S_P$ \citep{weinmann2017geometricfeatures,pauly2003multiscale}:

\begin{equation}
\begin{aligned}
    c_P &= \frac{\lambda_2-\lambda_3}{\lambda_1+\epsilon}, &
    \sigma_P &= \frac{\lambda_3}{\lambda_1+\lambda_2+\lambda_3}, \\
    \ell_P &= \frac{\lambda_1-\lambda_2}{\lambda_1}, &
    S_P &= \frac{\lambda_3}{\lambda_1}.
\end{aligned}
\label{eq:local_geometric_features}
\end{equation}

$c_P$, $\sigma_P$, $\ell_P$, and $S_P$ characterize local planarity, surface variation, linearity, and isotropy, respectively.

\paragraph{Robust normal estimation} High planarity indicates that the neighboring points are well approximated by a plane, making $\mathbf{e}_3$ a relatively reliable estimate of the surface normal, whereas lower planarity makes the local geometric normal less reliable. Multi-view normal estimates provide complementary surface-orientation cues, but their reliability varies across views. We therefore combine the two sources according to local planarity $c_P$ to obtain a more robust normal.

For each observing view $i\in\mathcal{V}_P$, we estimate depth using Depth Anything V2 \citep{yang2024depthanythingv2} and derive the corresponding normal map. The normal at the projection $\mathbf{p}_i$ of $P$ is transformed to world coordinates as $\mathbf{N}_{P,i}$. Because boundary projections are less reliable \citep{bae2024dsine}, each view is weighted by the distance between $\mathbf{p}_i$ and the principal point $\mathbf{q}_i$, then we normalize the weights over all views observing point $P$:

\begin{equation}
\begin{aligned}
    w_i^{\mathrm{proj}}
    &=
    \frac{1}{
        \left\|
        \mathbf{p}_i-\mathbf{q}_i
        \right\|_2+\delta
    },
    \\
    \widetilde{w}_i^{\mathrm{proj}}
    &=
    \frac{
        w_i^{\mathrm{proj}}
    }{
        \sum_{j\in\mathcal{V}_P}
        w_j^{\mathrm{proj}}
    }.
\end{aligned}
\label{eq:projection_weights}
\end{equation}

where $\delta$ prevents division by zero. Given the varying reliability across views, the normals $\mathbf{N}_{P,i}$ are fused using the normalized projection weights to obtain the multi-view normal:

\begin{equation}
    \overline{\mathbf{N}}_{\mathrm{view}}
    =
    \frac{
        \sum_{i\in\mathcal{V}_P}
        \widetilde{w}_i^{\mathrm{proj}}\mathbf{N}_{P,i}
    }{
        \left\|
        \sum_{i\in\mathcal{V}_P}
        \widetilde{w}_i^{\mathrm{proj}}\mathbf{N}_{P,i}
        \right\|_2
    }.
    \label{eq:multiview_normal}
\end{equation}

The fused multi-view normal $\overline{\mathbf{N}}_{\mathrm{view}}$ is used to resolve the orientation ambiguity of $\mathbf{e}_3$, after which both are fused according to local planarity $c_P$ to obtain the final robust normal $\mathbf{N}_P$:

\begin{equation}
\begin{aligned}
    \mathbf{N}_P^{\mathrm{str}}
    &=
    \operatorname{sign}
    \left(
    \overline{\mathbf{N}}_{\mathrm{view}}
    \cdot
    \mathbf{e}_3
    \right)
    \mathbf{e}_3,
    \\
    \mathbf{N}_P
    &=
    \frac{
        (1-c_P)\overline{\mathbf{N}}_{\mathrm{view}}
        +c_P\mathbf{N}_P^{\mathrm{str}}
    }{
        \left\|
        (1-c_P)\overline{\mathbf{N}}_{\mathrm{view}}
        +c_P\mathbf{N}_P^{\mathrm{str}}
        \right\|_2
    }.
\end{aligned}
\label{eq:robust_normal_fusion}
\end{equation}

\paragraph{Color feature construction}

To reduce exposure variations along long flight strips \citep{pastucha2022radiometric}, RGB colors are converted to CIELab \citep{cie2019lab}, yielding $\mathbf{C}_P=[L,a,b]^{\mathsf{T}}$.

\subsubsection{Supervoxel Clustering}
\label{subsubsec:supervoxel_clustering}

Based on the joint features described above, we perform structure-aware supervoxel clustering with curvature-guided seed initialization and a geometry-aware clustering distance.

\paragraph{Curvature-guided seed initialization} Since high-curvature regions typically contain edges or complex geometry, we allocate them denser seeds to better preserve geometric details.

As shown in Fig.~\ref{fig:pipeline}(a), the scene is voxelized, and the mean curvature $\overline{\sigma}_i$ is computed for each nonempty voxel $V_i$. Given $K_{\mathrm{total}}$ target seeds, let $r_i\in[0,1]$ denote the percentile rank of $\overline{\sigma}_i$. The number of seeds assigned to $V_i$ is:

\begin{equation}
    K_i=K_{\mathrm{total}}
    \frac{1+r_i}{\sum_j(1+r_j)}.
    \label{eq:seed_allocation}
\end{equation}

In implementation, the fractional allocations are converted to integer seed counts with residual adjustment to ensure $\sum_i K_i=K_{\mathrm{total}}$.

Seeds located in regions with abrupt geometric or color variations may adversely affect subsequent clustering. Each initial seed $S_{\mathrm{init}}$ is therefore moved to the point with the minimum combined normal and color gradient within its local candidate neighborhood, yielding the final clustering seed $S_k$. The combined gradient $G(P)$ is defined as:

\begin{equation}
    G(P)=\left\|\nabla\mathbf{N}_P\right\|_2^2
    +\left\|\nabla\mathbf{C}_P\right\|_2^2.
    \label{eq:seed_gradient}
\end{equation}

\paragraph{Structure-aware clustering distance} For each seed $S_k$, candidate points are restricted to a search radius of $2R_{\mathrm{grid}}$, where $R_{\mathrm{grid}}\approx\sqrt[3]{V/K_{\mathrm{total}}}$ and $V$ denotes the volume of the voxelized scene. Each candidate point $P$ is assigned to the seed minimizing $D(P,S_k)$, which combines spatial proximity, normal consistency, and color similarity. Spatial proximity promotes local connectivity, while normal and color differences distinguish nearby surfaces with different orientations or appearances:

\begin{figure}[!htbp]
    \centering
    \includegraphics[width=0.50\columnwidth]{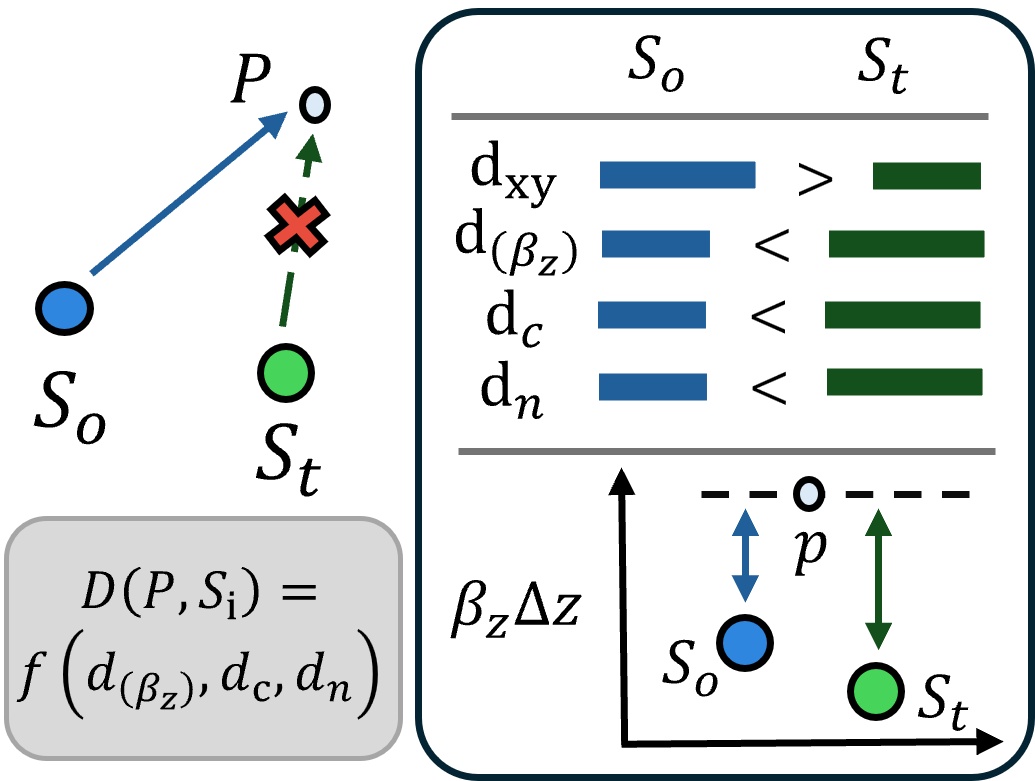}
    \caption{Anisotropic spatial distance with vertical penalty $\beta_z$ for distinguishing adjacent surfaces at different elevations.}
    \label{fig:spatial_distance}
\end{figure}

\begin{equation}
\begin{aligned}
    D(P,S_k)
    &=
    \sqrt{
        d_z(P,S_k)^2
        +d_n(P,S_k)^2
        +d_c(P,S_k)^2
    },\\
    d_z(P,S_k)
    &=
    \sqrt{
        \left\|P_{xy}-S_{k,xy}\right\|_2^2
        +\beta_z\left\|P_z-S_{k,z}\right\|_2^2
    },\\
    d_n(P,S_k)
    &=
    1-\left|\mathbf{N}_P\cdot\mathbf{N}_{S_k}\right|,\\
    d_c(P,S_k)
    &=
    \left\|\mathbf{C}_P-\mathbf{C}_{S_k}\right\|_{\mathrm{Lab}}.
\end{aligned}
\label{eq:clustering_distance}
\end{equation}

where $d_z$, $d_n$, and $d_c$ denote the spatial, normal, and color terms, respectively. Each is normalized before aggregation to balance scale differences.

As illustrated in Fig.~\ref{fig:spatial_distance}, the vertical penalty $\beta_z$ makes clustering more sensitive to elevation variations. This is particularly useful for near-nadir aerial imagery because surfaces at different elevations, such as roofs and ground, tree canopies, or overpasses, may overlap in horizontal projection.

\subsubsection{Sub-region Organization and Boundary Optimization}
\label{subsubsec:subregion_boundary}

Following supervoxel clustering, RAG-based region merging and training-view assignment prepare the sub-regions for independent optimization. After all sub-regions have been independently optimized, boundary refinement is performed as the final optimization stage to improve consistency across adjacent sub-regions.

\paragraph{RAG-based region merging} Although supervoxel clustering preserves local structure, continuous elements may still be split into adjacent clustered regions by local geometric variations. To reduce this over-segmentation, we construct a region adjacency graph (RAG), where each region is a node and edges connect only spatially adjacent regions, avoiding exhaustive pairwise comparisons.

For each adjacent region pair $(A,B)$ connected by an edge in the RAG, let $\boldsymbol{\mu}_A$ and $\boldsymbol{\mu}_B$ and $\mathbf{N}_A,\mathbf{N}_B$ denote their centroids and average normals, respectively, with $\mathbf{d}=\boldsymbol{\mu}_A-\boldsymbol{\mu}_B$. They are merged when

\begin{equation}
    \left(\mathbf{N}_A\times\mathbf{d}\right)\cdot
    \left(\mathbf{N}_B\times\mathbf{d}\right)>0
    \label{eq:rag_plane_relation}
\end{equation}

and the angle between $\mathbf{N}_A$ and $\mathbf{N}_B$ is below $10^\circ$, indicating that the regions likely belong to the same local structure \citep{rabbani2006smoothness,maalek2018robust}.

\paragraph{Training-view assignment} Each sub-region is optimized using only the views that provide sufficient coverage. For a sub-region $A$, its SfM points are projected into each candidate view and views satisfying the visibility threshold are assigned to $A$. The convex hull of the projected points defines a binary mask $M$, within which the rendering loss is evaluated, limiting updates across different partitions.

\paragraph{Boundary-consistency constraints} Independently optimized sub-regions may be inconsistent along shared boundaries. Each boundary is therefore expanded by 20\%, and the intersection between adjacent expanded regions defines the overlap band $\Omega_{\mathrm{ov}}$. Commonly visible views are selected using the SfM co-visibility relationships within this band. After independent optimization, Gaussians outside $\Omega_{\mathrm{ov}}$ are fixed, and only those inside are refined using color and depth consistency:

\begin{equation}
    \mathcal{L}_{\mathrm{ov}}
    =
    \left\|I_A(\Omega_{\mathrm{ov}})-I_B(\Omega_{\mathrm{ov}})\right\|_1
    +
    \left\|D_A(\Omega_{\mathrm{ov}})-D_B(\Omega_{\mathrm{ov}})\right\|_1.
    \label{eq:overlap_consistency}
\end{equation}

Meanwhile, because adjacent sub-regions may introduce redundant Gaussians in the overlap band, we apply opacity sparsity regularization:

\begin{equation}
    \mathcal{L}_{\mathrm{sparse}}^{A,B}
    =
    \frac{1}{2}
    \left(
    \frac{1}{\left|\mathcal{G}_A^{\mathrm{ov}}\right|}
    \sum_{g\in\mathcal{G}_A^{\mathrm{ov}}}\alpha_g
    +
    \frac{1}{\left|\mathcal{G}_B^{\mathrm{ov}}\right|}
    \sum_{g\in\mathcal{G}_B^{\mathrm{ov}}}\alpha_g
    \right).
    \label{eq:overlap_opacity_sparsity}
\end{equation}

The final boundary-consistency loss combines color and depth consistency with opacity sparsity regularization:

\begin{equation}
    \mathcal{L}_{\mathrm{boundary}}
    =
    \mathcal{L}_{\mathrm{ov}}
    +
    \lambda_{\mathrm{sp}}
    \mathcal{L}_{\mathrm{sparse}}^{A,B}.
    \label{eq:boundary_loss}
\end{equation}

The $\lambda_{\mathrm{sp}}$ controls the weight of the opacity sparsity regularization.

\subsection{Neighborhood-Aware Gaussian Organization}
\label{subsec:topology_organization}

Independent optimization of Gaussian primitives lacks explicit constraints on their neighborhood-level geometric organization, causing the local arrangement of neighboring Gaussians to become increasingly unstable during training. As shown in Fig.~\ref{fig:pipeline}(b), we therefore introduce neighborhood-aware Gaussian organization, which maintains a geometry-aware graph to encode local geometric relationships and impose normal-consistency and tangential-distribution constraints over graph neighborhoods to improve local consistency. During Gaussian densification, the graph is updated accordingly to remain aligned with the evolving primitive distribution.

\subsubsection{Geometry-Aware Neighborhood Graph Initialization}
\label{subsubsec:topology_initialization}

We model Gaussian adjacency as an undirected neighborhood graph $\mathcal{G}=(\mathcal{V},\mathcal{E})$, where each vertex represents a 3D Gaussian primitive. Candidate edges are first restricted by spatial proximity and then pruned by geometric consistency. Each Gaussian initialized from SfM point inherits the corresponding geometric attributes, including curvature and planarity. Its orientation is initialized by aligning the shortest principal axis of $\boldsymbol{\Sigma}_i$ with the robust normal estimated in Section~\ref{subsubsec:joint_features}, while the Gaussian normal $\mathbf{n}_i$ is differentiably derived from this axis throughout training.

For each primitive $i$, its $K$ nearest Gaussians within radius $r_{\mathrm{max}}$ form the candidate neighborhood:

\begin{equation}
    \mathcal{N}_i =
    \left\{
    j \in \mathrm{KNN}_K(i)
    \,\middle|\,
    \left\|
    \boldsymbol{\mu}_i-\boldsymbol{\mu}_j
    \right\|_2
    \leq r_{\mathrm{max}}
    \right\}.
    \label{eq:topology_candidate}
\end{equation}

Candidate neighbors are further evaluated by a geometric consistency score $s_{ij}$ combining normal and curvature similarity:

\begin{equation}
    s_{ij}
    =
    \frac{1}{2}
    \left[
    \left|\mathbf{n}_i^{\mathsf{T}}\mathbf{n}_j\right|
    +1
    -3\left|\sigma_i-\sigma_j\right|
    \right].
    \label{eq:topology_consistency}
\end{equation}

The top $50\%$ of candidates ranked by $s_{ij}$ are retained as $\widetilde{\mathcal{N}}_i$, and the initial edge set is defined as

\begin{equation}
    \mathcal{E}^{(0)}
    =
    \left\{
    (i,j)
    \;\middle|\;
    j\in\widetilde{\mathcal{N}}_i
    \;\text{or}\;
    i\in\widetilde{\mathcal{N}}_j
    \right\}.
    \label{eq:initial_topology_edges}
\end{equation}

The resulting graph organizes spatially close and geometrically consistent Gaussians into local neighborhoods for subsequent geometric optimization. Since robust normals and curvature are precomputed during scene partitioning, graph initialization introduces no additional feature-extraction cost.

\subsubsection{Neighborhood Geometric Constraints}
\label{subsubsec:topology_constraints}

Within each graph neighborhood, we jointly constrain surface orientation and primitive distribution: normal consistency preserves local orientation continuity, while tangential consistency suppresses redundant stacking along the surface normal.

\paragraph{Neighborhood normal-consistency constraint} Neighboring Gaussians on the local surface and connected in the graph should have similar orientations. The normal-consistency loss is defined as

\begin{equation}
    \mathcal{L}_{\mathrm{nor}}
    =
    \sum_{i\in\mathcal{V}}
    \sum_{j\in\mathcal{N}(i)}
    \left(
    1-\left|\mathbf{n}_i\cdot\mathbf{n}_j\right|
    \right).
    \label{eq:normal_consistency}
\end{equation}

\paragraph{Neighborhood tangential-distribution constraint} For neighboring Gaussians on the same local surface, the center displacement should lie primarily in the tangent plane. We therefore penalize the Gaussian along the normal direction $\mathbf{n}_{\mathrm{avg}}=(\mathbf{n}_i+\mathbf{n}_j)/2$, suppressing redundant stacking, with $\mathbf{v}_{ij}=\boldsymbol{\mu}_j-\boldsymbol{\mu}_i$:

\begin{equation}
    \mathcal{L}_{\mathrm{tan}}
    =
    \sum_{i\in\mathcal{V}}
    \sum_{j\in\mathcal{N}(i)}
    \left(
    \mathbf{v}_{ij}\cdot\mathbf{n}_{\mathrm{avg}}
    \right)^2.
    \label{eq:tangential_distribution}
\end{equation}

\subsubsection{Gradient Modulation and Dynamic Graph Update}
\label{subsubsec:gradient_graph_update}

To maintain neighborhood organization during optimization, neighborhood-average normals are used to suppress geometry-inconsistent gradient updates, while the graph is synchronized with the evolving Gaussian distribution by updating the structural attributes and neighborhood connections of newly split or cloned primitives.

\paragraph{Gradient modulation} For primitive $i$, $\hat{\mathbf{n}}_i$ is the normal predicted from a provisional parameter update, and $\bar{\mathbf{n}}_{\mathcal{N}_(i)}$ is the neighborhood-average normal. Their deviation defines the modulation weight:

\begin{equation}
    w_i^n
    =
    \exp
    \left(
    -\left\|
    \hat{\mathbf{n}}_i
    -
    \bar{\mathbf{n}}_{\mathcal{N}(i)}
    \right\|
    \right).
    \label{eq:normal_guided_gradient_weight}
\end{equation}

The provisional update is used only to estimate $\hat{\mathbf{n}}_i$ and is not committed; the resulting $w_i^n$ is then used to modulate the gradient in the actual parameter update. Thus, neighborhood-consistent updates retain larger weights, whereas inconsistent updates are suppressed, reducing updates that are inconsistent with the neighborhood geometry and maintaining local geometric consistency during optimization.

\paragraph{Dynamic graph update} When a primitive $G_i$ is split or cloned, the new primitives initially inherit the parent's normal, curvature, and planarity, while the parent's neighbors are used as candidate neighbors for the new primitives. The candidate connections are then re-evaluated using the same spatial and geometric criteria as in Section~\ref{subsubsec:topology_initialization}, with the top 50\% retained.

Their inherited structural attributes are then refined by neighborhood-based interpolation to match the updated local geometry.

\subsection{Adaptive Surface Regularization}
\label{subsec:adaptive_surface_regularization}

Large-scale aerial scenes typically contain diverse scene elements with substantially different geometric and structural complexity. Structured regions, such as building roofs and roads, are generally well approximated by local planes and thus exhibit higher local planarity, whereas unstructured regions, such as vegetation and gravel surfaces, contain more pronounced local geometric variations and exhibit lower local planarity. A uniform surface regularization strength cannot simultaneously provide sufficient geometric constraints in structured regions and preserve plausible local variations in unstructured regions. To address this issue, we adapt surface regularization strength based on local planarity without requiring explicit semantic labels, as shown in Fig.~\ref{fig:pipeline}(c).

\subsubsection{Opacity-Binarization Regularization}
\label{subsubsec:opacity_binarization}

Opacity-binarization regularization encourages Gaussian opacities toward 0 or 1, thereby penalizing redundant accumulation of Gaussians with intermediate opacity values ($\alpha\approx 0.5$). Together with the neighborhood tangential distribution constraint, it reduces blurring at surface boundaries. However, applying a uniform binarization constraint across the scene may over-constrain valid geometric variations in unstructured regions. This limitation calls for region-adaptive weighting of the opacity-binarization constraint. Specifically, the opacity-binarization loss is defined as

\begin{equation}
    \mathcal{L}_{\mathrm{bin}}
    =
    \frac{1}{|\mathcal{S}|}
    \sum_{g\in\mathcal{S}}
    \mathbb{M}_{\mathrm{adapt}}(g)\,
    \Phi(\alpha_g),
    \label{eq:opacity_binarization}
\end{equation}

\begin{equation}
    \Phi(\alpha)
    =
    -\left[
    \alpha\log(\alpha+\epsilon)
    +(1-\alpha)\log(1-\alpha+\epsilon)
    \right].
    \label{eq:opacity_entropy}
\end{equation}

The set $\mathcal{S}$ contains the Gaussians contributing to surface rendering in the current view, and $\alpha_g$ is the opacity of Gaussian $g$. The adaptive weight $\mathbb{M}_{\mathrm{adapt}}(g)$ is determined by the local planarity $c_p(g)$ and is defined as:

\begin{equation}
    \mathbb{M}_{\mathrm{adapt}}(g)
    =
    \frac{1+c_p(g)}
    {
        \frac{1}{|\mathcal{S}|}
        \sum_{g\in\mathcal{S}}\left(1+c_p(g)\right)
    }.
    \label{eq:adaptive_opacity_weight}
\end{equation}

Higher-planarity structured regions are assigned larger weights, whereas lower-planarity unstructured regions receive smaller weights.

\subsubsection{Depth-Variance Regularization}
\label{subsubsec:depth_variance}

Depth-variance regularization constrains the depth distribution of Gaussians contributing to each pixel, encouraging their depths to concentrate around a consistent value. This reduces ambiguity in surface depth estimation and alleviates surface blurring and boundary expansion. However, a uniform depth-variance constraint cannot simultaneously concentrate the depth distribution in structured regions and preserve plausible depth variations in unstructured regions. This limitation requires the regularization strength to be adapted across regions. Specifically, the depth-variance loss is defined as

\begin{equation}
    \mathcal{L}_{\mathrm{var}}
    =
    \frac{1}{|\mathcal{U}|}
    \sum_{u\in\mathcal{U}}
    \mathbf{W}_{\mathrm{adapt}}(u)\,
    \operatorname{Var}(d)_u.
    \label{eq:depth_variance}
\end{equation}

The adaptive weight $\mathbf{W}_{\mathrm{adapt}}(u)$ is computed from the planarity response $R_u$ at pixel $u$ and is defined as:

\begin{equation}
    \mathbf{W}_{\mathrm{adapt}}(u)
    =
    \frac{1+R_u}
    {
        \frac{1}{|\mathcal{U}|}
        \sum_{u\in\mathcal{U}}(1+R_u)
    },
    \label{eq:adaptive_depth_weight}
\end{equation}

\begin{equation}
    R_u
    =
    \frac{
        \sum_{g\in\mathcal{G}(u)}
        \omega_g(u)c_p(g)
    }{
        \sum_{g\in\mathcal{G}(u)}
        \omega_g(u)+\epsilon
    }.
    \label{eq:pixel_planarity_response}
\end{equation}

In these equations, $\mathcal{U}$ denotes the set of surface pixels in the current view, and $\operatorname{Var}(d)_u$ is the variance of the depth distribution at pixel $u$. The set $\mathcal{G}(u)$ contains the Gaussians that contribute to pixel $u$, while $\omega_g(u)$ denotes its $\alpha$-compositing contribution as defined in Section~\ref{subsec:3dgs}.

Pixels with larger planarity responses, typically corresponding to structured regions, receive larger weights, whereas pixels with smaller responses, typically associated with unstructured regions, receive smaller weights to avoid suppressing plausible depth variations.

\subsection{Overall Optimization Objective}
\label{subsec:overall_objective}

The loss terms are activated progressively to avoid imposing geometric regularization before the primitives establish a reasonable initial spatial distribution. Training starts with the rendering loss, the neighborhood geometric constraints are then activated to maintain local geometric organization during subsequent optimization, followed by adaptive surface regularization. During post-optimization boundary refinement, the boundary-consistency loss is activated only within overlapping areas of adjacent sub-regions. The overall objective is

\begin{equation}
\begin{aligned}
    \mathcal{L}
    &=
    \mathcal{L}_{\mathrm{render}}
    +\lambda_{\mathrm{nbr}}
    \left(
        \mathcal{L}_{\mathrm{nor}}
        +\mathcal{L}_{\mathrm{tan}}
    \right) \\
    &\quad
    +\lambda_{\mathrm{surf}}
    \left(
        \mathcal{L}_{\mathrm{bin}}
        +\mathcal{L}_{\mathrm{var}}
    \right)
    +\lambda_{\mathrm{bd}}
    \mathcal{L}_{\mathrm{boundary}}.
\end{aligned}
\label{eq:overall_objective}
\end{equation}

where $\lambda_{\mathrm{nbr}}$, $\lambda_{\mathrm{surf}}$, and $\lambda_{\mathrm{bd}}$ weight the neighborhood constraints, adaptive surface regularization, and boundary refinement, respectively. $\mathcal{L}{\mathrm{render}}$ and $\mathcal{L}{\mathrm{boundary}}$ are defined in Eqs.~\ref{eq:rendering_loss} and~\ref{eq:boundary_loss}, respectively. Specific activation timings and weight schedules are provided in the implementation details in Section~\ref{subsec:experimental_setup}.

\begin{table}[t]
\centering
\caption{Statistics of the datasets and selected scenes.}
\label{tab:dataset_statistics}
\small
\setlength{\tabcolsep}{4.5pt}
\renewcommand{\arraystretch}{1.08}
\begin{tabular}{@{}llrcr@{}}
\toprule
Dataset & Scene & Images & Resolution & Area ($\mathrm{km}^2$) \\
\midrule
\multirow{4}{*}{GauU-Scene}
& Lower Campus & 670  & $5474 \times 3643$ & 1.14 \\
& Upper Campus & 713  & $5469 \times 3638$ & 1.01 \\
& LFLS         & 1077 & $5468 \times 3638$ & 1.86 \\
& SZIIT        & 1215 & $5466 \times 3634$ & 1.85 \\
\addlinespace
\multirow{1}{*}{AIRLY}
& AIRLY        & 602  & $6000 \times 3999$ & 0.325 \\
\addlinespace
\multirow{2}{*}{UrbanScene3D}
& Residence    & 2582 & $5472 \times 3648$ & \textemdash \\
& Sci-Art      & 3019 & $4864 \times 3648$ & \textemdash \\
\addlinespace
\multirow{2}{*}{Mill19}
& Building     & 1940 & $4608 \times 3456$ & 0.125 \\
& Rubble       & 1678 & $4608 \times 3456$ & \textemdash \\
\bottomrule
\end{tabular}
\end{table}

\section{Experiments}
\label{sec:experiment}

\begin{figure*}[t]
    \centering
    \includegraphics[width=\textwidth]{
        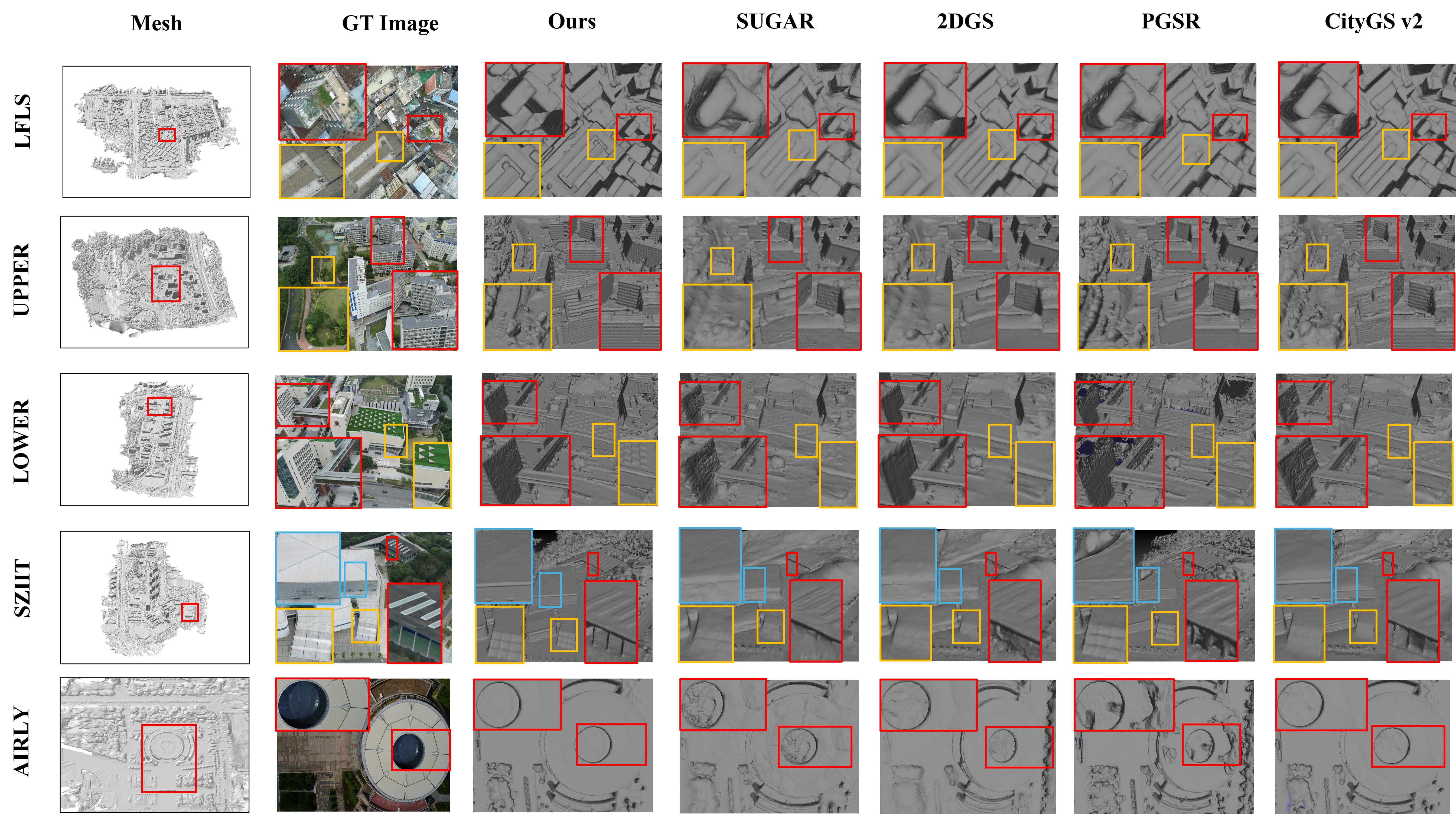
    }
    \caption{Local surface reconstruction details of different methods on GauU-Scene \citep{xiong2024gauuscene} and AIRLY \citep{yao2026arsgaussian}. Representative regions are shown to compare building outlines, road boundaries, vegetation undulations, and local geometric completeness.}
    \label{fig:gauu_airly_local_details}
\end{figure*}

\subsection{Experimental Setup}
\label{subsec:experimental_setup}

\paragraph{Dataset Preparation} We evaluate STARS-GS on four large-scale reconstruction datasets with complementary scene characteristics and evaluation purposes: GauU-Scene \citep{xiong2024gauuscene} and AIR-LONGYAN (AIRLY) \citep{yao2026arsgaussian} are used for the primary quantitative and qualitative evaluation, while UrbanScene3D \citep{lin2022urbanscene3d} and Mill19 \citep{turki2022meganerf} are used for supplementary qualitative comparisons.

GauU-Scene and AIRLY are selected as the primary quantitative evaluation datasets because they both contain high-resolution multi-view aerial images together with reference LiDAR point clouds, as well as diverse scene elements, including buildings, roads, and vegetation, enabling quantitative geometric evaluation under large-scale conditions. In addition, AIRLY contains multiple types of scene elements that are more concentrated in the central part of the scene and closely adjacent in space, resulting in more pronounced local variations in geometric morphology and structural complexity. This characteristic makes AIRLY particularly suitable for detailed ablation visualization and region-wise analysis.

\begin{figure}[!t]
    \centering
    \includegraphics[width=\linewidth]{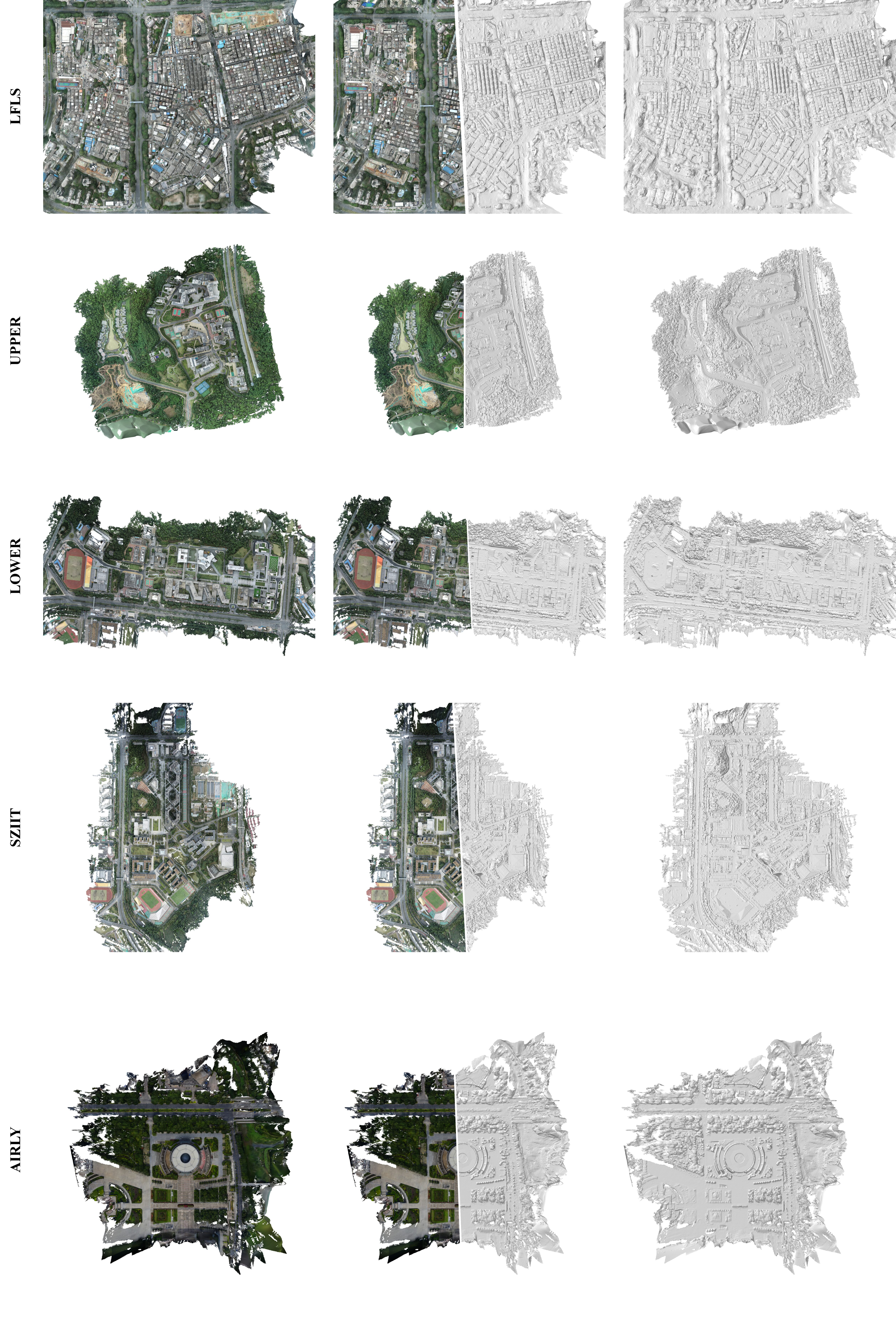}
    \caption{Complete surface meshes reconstructed by STARS-GS on GauU-Scene \citep{xiong2024gauuscene} and AIRLY \citep{yao2026arsgaussian}. The overall textured meshes and representative local details illustrate surface completeness and structural continuity in large-scale scenes.}
    \label{fig:gauu_airly_complete_mesh}
\end{figure}

\begin{figure}[!t]
    \centering
    \includegraphics[width=\linewidth]{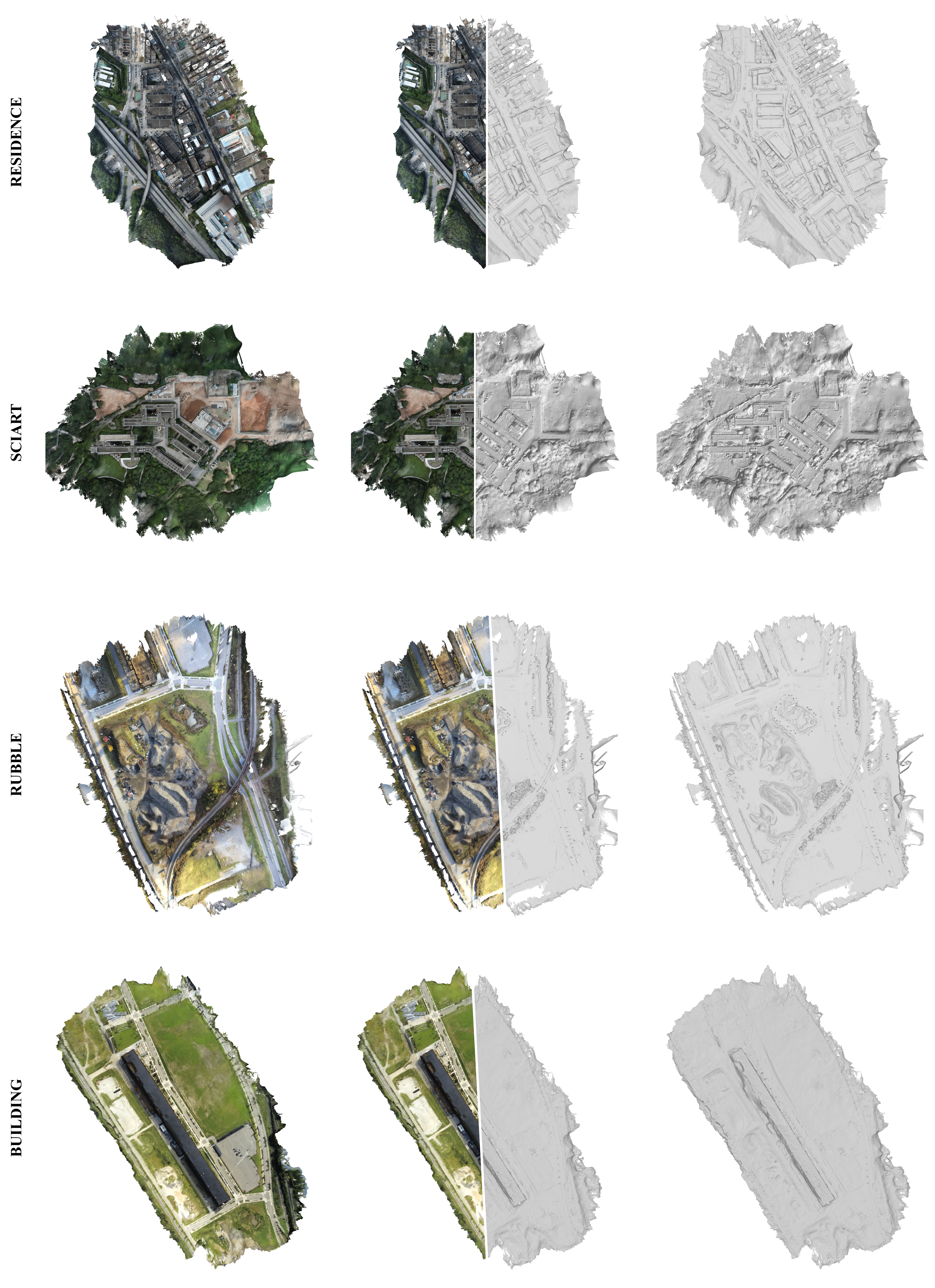}
    \caption{Complete surface reconstruction results of STARS-GS on  UrbanScene3D \citep{lin2022urbanscene3d} and Mill19 \citep{turki2022meganerf}. The overall meshes and local details are shown for dense building scenes and irregular scenes.}
    \label{fig:urbanscene3d_mill19_complete_mesh}
\end{figure}

Although UrbanScene3D and Mill19 do not provide reference point clouds for quantitative geometric evaluation thus cannot support the primary quantitative evaluation in this study, both are widely used public datasets in large-scale reconstruction studies, better facilitating comparison with existing large-scale surface reconstruction methods. In addition, the selected UrbanScene3D scenes are dominated by densely distributed buildings, with relatively continuous surfaces and well-defined geometric boundaries. By contrast, the selected Mill19 scenes mainly contain bare ground and vegetation, where local surface variations are more pronounced, image textures are weaker, and geometric structures and boundaries are more irregular. These different scene characteristics further extend the diversity of the evaluation settings. Therefore, UrbanScene3D and Mill19 are included for supplementary qualitative evaluation to examine the robustness and consistency of STARS-GS across additional datasets with different scene characteristics.

Detailed statistics of all datasets and selected scenes, including the number of images, image resolution, and scene area, are summarized in Table~\ref{tab:dataset_statistics}. Following prior work, all input images are downsampled by a factor of four before training \citep{turki2022meganerf,lin2024vastgaussian}.

\paragraph{Comparison Methods}
We compare STARS-GS with representative Gaussian-based surface reconstruction methods that provide publicly available implementations and can be reproduced under a unified experimental protocol, including SuGaR \citep{guedon2024sugar}, 2DGS \citep{huang2024twodgs}, PGSR \citep{chen2025pgsr}, and CityGaussianV2 \citep{liu2025citygaussianv2}. All methods are evaluated using their official implementations and recommended configurations without additional scene-specific tuning. To ensure consistent geometric evaluation, the reconstructed results of all methods are converted into meshes using the 2DGS mesh-extraction pipeline \citep{huang2024twodgs}. Median depth is used for TSDF integration, and identical mesh-extraction parameters are adopted for all methods within each scene.

However, not all relevant methods can be evaluated under a unified full-scene geometric protocol. Some Gaussian-based surface reconstruction methods are difficult to apply directly to the large scenes in our main evaluation, whereas several large-scale 3DGS methods focus on rendering and lack an explicit surface-reconstruction pipeline. We therefore conduct task-specific supplementary comparisons on AIRLY. STARS-GS is compared with GSDF \citep{yu2024gsdf}, RaDe-GS \citep{zhang2026radegs}, and Trim2DGS \citep{fan2024trimgs} for surface reconstruction, and with BlockGaussian \citep{wu2025blockgaussian} and VastGaussian \citep{lin2024vastgaussian} for novel view synthesis.

\paragraph{Implementation Details} All experiments are conducted on the same computing platform equipped with two NVIDIA RTX A6000 GPUs. For STARS-GS, the minimum observation threshold in sparse point-cloud preprocessing, $\tau_{\mathrm{track}}$, is set to 5. Local geometric feature extraction and neighborhood graph construction both use $k=20$ nearest-neighbor search with a maximum search radius of $r_{\max}=0.005d_{\mathrm{scene}}$, where $d_{\mathrm{scene}}$ denotes the diagonal length of the scene bounding box. In structure-aware scene partitioning, the global target number of seeds $K_{\mathrm{total}}$ is set to 12, and the vertical penalty factor is set to $\beta_z=2.0$.

Each sub-region is optimized for 30,000 iterations and Gaussian densification is performed every 250 iterations, after which the geometry-aware neighborhood graph is updated accordingly. The neighborhood geometric constraint $\mathcal{L}_{\mathrm{nbr}}$ is activated after 7000 iterations. Its weight $\lambda_{\mathrm{nbr}}$ is then linearly increased from $0.5\lambda_{\mathrm{nbr}}^{\mathrm{max}}$ to $\lambda_{\mathrm{nbr}}^{\mathrm{max}}$ over the next 15000 iterations to avoid imposing strong neighborhood constraints before the primitives establish a meaningful initial spatial distribution, with $\lambda_{\mathrm{nbr}}^{\mathrm{max}} = 0.01$ in the default setting. The adaptive surface regularization term $\mathcal{L}_{\mathrm{surf}}$ is activated after 15000 iterations, and its weight $\lambda_{\mathrm{surf}}$ is linearly increased from $0.5\lambda_{\mathrm{surf}}^{\mathrm{max}}$ to $\lambda_{\mathrm{surf}}^{\mathrm{max}}$ over the next 7000 iterations, with default $\lambda_{\mathrm{surf}}^{\mathrm{max}} = 0.01$ .

After training each sub-region, the overlapping areas between adjacent sub-regions are refined for an additional 3,000 iterations, while Gaussian parameters outside the overlapping areas are kept fixed. The boundary-consistency loss $\mathcal{L}_{\mathrm{boundary}}$ is activated only in this stage, with $\lambda_{\mathrm{bd}}=0.01$. The weight of the opacity sparsity regularization term is set to $\lambda_{\mathrm{sp}}=0.01$. Unless otherwise specified, the remaining parameters follow the default configuration of the vanilla 3DGS \citep{kerbl2023gaussians}.

\paragraph{Evaluation Metrics} We evaluate all methods in terms of geometric reconstruction accuracy and novel view synthesis quality.

Geometric reconstruction accuracy is the primary criterion for large-scale surface reconstruction. Because different methods may produce different geometric representations, including explicit meshes, depth-derived point clouds, and Gaussian primitives, direct comparison of their raw outputs may compromise the fairness of the evaluation. We consequently follow the unified geometric evaluation protocol introduced in CityGaussianV2 \citep{liu2025citygaussianv2}. Specifically, a surface mesh is first extracted from the result of each method using the 2DGS mesh-extraction procedure \citep{huang2024twodgs}, with a voxel size of 1 m and an SDF truncation distance of 4 m. The same mesh-extraction configuration is used for all compared methods within each scene. The same number of points is then uniformly sampled from each mesh surface, and both the sampled points and LiDAR ground-truth point cloud are cropped to the common valid evaluation region. Following CityGaussianV2, the distance threshold $\tau$ is determined from the nearest-neighbor distance statistics of the downsampled reference point cloud and falls within the range of 0.3-0.6 m. Based on the processed reconstruction and ground-truth point clouds, Precision, Recall, and F1-score are computed as the geometric evaluation metrics. Precision and Recall measure reconstruction accuracy and completeness, respectively, while F1-score provides their harmonic mean.

Since geometry and appearance in 3DGS are represented by the same Gaussian primitives, geometric constraints may also affect rendering quality. We therefore additionally evaluate novel view synthesis using PSNR, SSIM \citep{wang2004ssim}, and LPIPS \citep{zhang2018lpips} to measure image rendering quality and perceptual consistency.

\subsection{Experimental Results}
\label{subsec:experimental_results}

\subsubsection{Surface Reconstruction}
\label{subsubsec:surface_reconstruction}

\begin{table*}[t]
\centering
\caption{Quantitative surface reconstruction results on GauU-Scene \citep{xiong2024gauuscene} and AIRLY \citep{yao2026arsgaussian}. We report Precision, Recall, F1-score, together with the average training time and average number of Gaussian primitives. The best and second-best results are shown in \textbf{bold} and \underline{underlined}, respectively.}
\label{tab:surface_reconstruction_main}
\scriptsize
\setlength{\tabcolsep}{2.5pt}
\resizebox{\textwidth}{!}{
\begin{tabular}{l*{17}{c}}
\toprule
\multirow{2}{*}{Method}
& \multicolumn{3}{c}{Lower Campus}
& \multicolumn{3}{c}{Upper Campus}
& \multicolumn{3}{c}{LFLS}
& \multicolumn{3}{c}{SZIIT}
& \multicolumn{3}{c}{AIRLY}
& \multirow{2}{*}{\shortstack{Time\\(h) $\downarrow$}}
& \multirow{2}{*}{\shortstack{Gaussians\\(M) $\downarrow$}} \\
\cmidrule(lr){2-4}
\cmidrule(lr){5-7}
\cmidrule(lr){8-10}
\cmidrule(lr){11-13}
\cmidrule(lr){14-16}
& P $\uparrow$ & R $\uparrow$ & F1 $\uparrow$
& P $\uparrow$ & R $\uparrow$ & F1 $\uparrow$
& P $\uparrow$ & R $\uparrow$ & F1 $\uparrow$
& P $\uparrow$ & R $\uparrow$ & F1 $\uparrow$
& P $\uparrow$ & R $\uparrow$ & F1 $\uparrow$
& & \\
\midrule
SuGaR
& 0.412 & 0.348 & 0.377
& 0.396 & 0.315 & 0.351
& 0.381 & 0.367 & 0.374
& 0.365 & 0.274 & 0.313
& 0.314 & 0.295 & 0.304
& 4.20 & \textbf{15.44} \\
2DGS
& 0.592 & 0.601 & 0.596
& 0.574 & 0.530 & 0.551
& 0.558 & 0.485 & 0.519
& 0.543 & 0.596 & 0.568
& 0.511 & 0.463 & 0.486
& \textbf{2.42} & \underline{22.06} \\
PGSR
& \underline{0.682} & 0.568 & 0.620
& 0.629 & 0.661 & 0.645
& 0.634 & 0.596 & 0.614
& \underline{0.644} & 0.575 & 0.608
& 0.501 & 0.577 & 0.536
& 5.60 & 28.14 \\
CityGaussianV2
& 0.673 & \underline{0.631} & \underline{0.651}
& \textbf{0.701} & \underline{0.675} & \underline{0.688}
& \underline{0.652} & \underline{0.638} & \underline{0.645}
& 0.641 & \underline{0.603} & \underline{0.621}
& \underline{0.610} & \underline{0.582} & \underline{0.596}
& 3.96 & 59.30 \\
STARS-GS
& \textbf{0.708} & \textbf{0.694} & \textbf{0.701}
& \underline{0.695} & \textbf{0.712} & \textbf{0.703}
& \textbf{0.686} & \textbf{0.690} & \textbf{0.688}
& \textbf{0.673} & \textbf{0.689} & \textbf{0.681}
& \textbf{0.723} & \textbf{0.716} & \textbf{0.719}
& \underline{3.60} & 24.20 \\
\bottomrule
\end{tabular}
}
\end{table*}

\paragraph{Results on GauU-Scene and AIRLY} Table~\ref{tab:surface_reconstruction_main} reports the quantitative surface reconstruction results on GauU-Scene \citep{xiong2024gauuscene} and AIRLY \citep{yao2026arsgaussian}. STARS-GS achieves an average F1-score of 0.693 on GauU-Scene, improving over the second-best CityGaussianV2 (0.651) by approximately 6.5\%, and reaches 0.719 on AIRLY, outperforming the second-best result (0.596) by approximately 20.6\%. STARS-GS obtains the highest F1-score on all five test scenes, demonstrating that STARS-GS consistently improved geometric reconstruction accuracy and completeness. Despite STARS-GS introducing structure-aware scene partitioning, neighborhood geometric constraints, and adaptive surface regularization, its average training time is 3.60 h, lower than SuGaR, PGSR, and CityGaussianV2. The average number of Gaussian primitives is 24.20 M, also lower than PGSR and CityGaussianV2. These results show that STARS-GS improves surface reconstruction quality while maintaining competitive computational and representation efficiency.

Fig.~\ref{fig:gauu_airly_local_details} presents local surface reconstruction results of different methods in representative regions of GauU-Scene and AIRLY, illustrating how the quantitative metric differences in Table~\ref{tab:surface_reconstruction_main} are reflected in specific geometric structures. SuGaR and 2DGS recover the main surface structures of scene elements, but evident over-smoothing and geometric drift remain around building outlines, road boundaries, and vegetation regions, making it difficult to preserve fine local structures. These qualitative observations are consistent with their relatively lower Precision and Recall. PGSR preserves relatively rich local details and structural variations, but noticeable holes and contour omissions remain in the reconstructed surfaces. These missing structures explain its relatively high Precision but lower Recall. CityGaussianV2 reconstructs regular building and road structures relatively accurately, but geometric undulations on vegetation surfaces and some fine structures remain oversmoothed, and local missing regions still limit complete recovery of the real surface.

In comparison, the local reconstruction results show that STARS-GS more accurately recovers building outlines, road boundaries, vegetation undulations, and local geometric details, while reducing geometric drift and surface incompleteness. Fig.~\ref{fig:gauu_airly_complete_mesh} further presents the full-scene surface meshes generated by STARS-GS on GauU-Scene and AIRLY. Across these large-scale aerial scenes, STARS-GS consistently reconstructs the overall scene geometry,, without obvious large holes or structural omissions in the resulting meshes. Together, the local and full-scene results demonstrate that STARS-GS consistently improves both geometric accuracy and surface completeness, leading to better overall reconstruction performance.

\begin{figure*}[t]
    \centering
    \includegraphics[width=\textwidth]{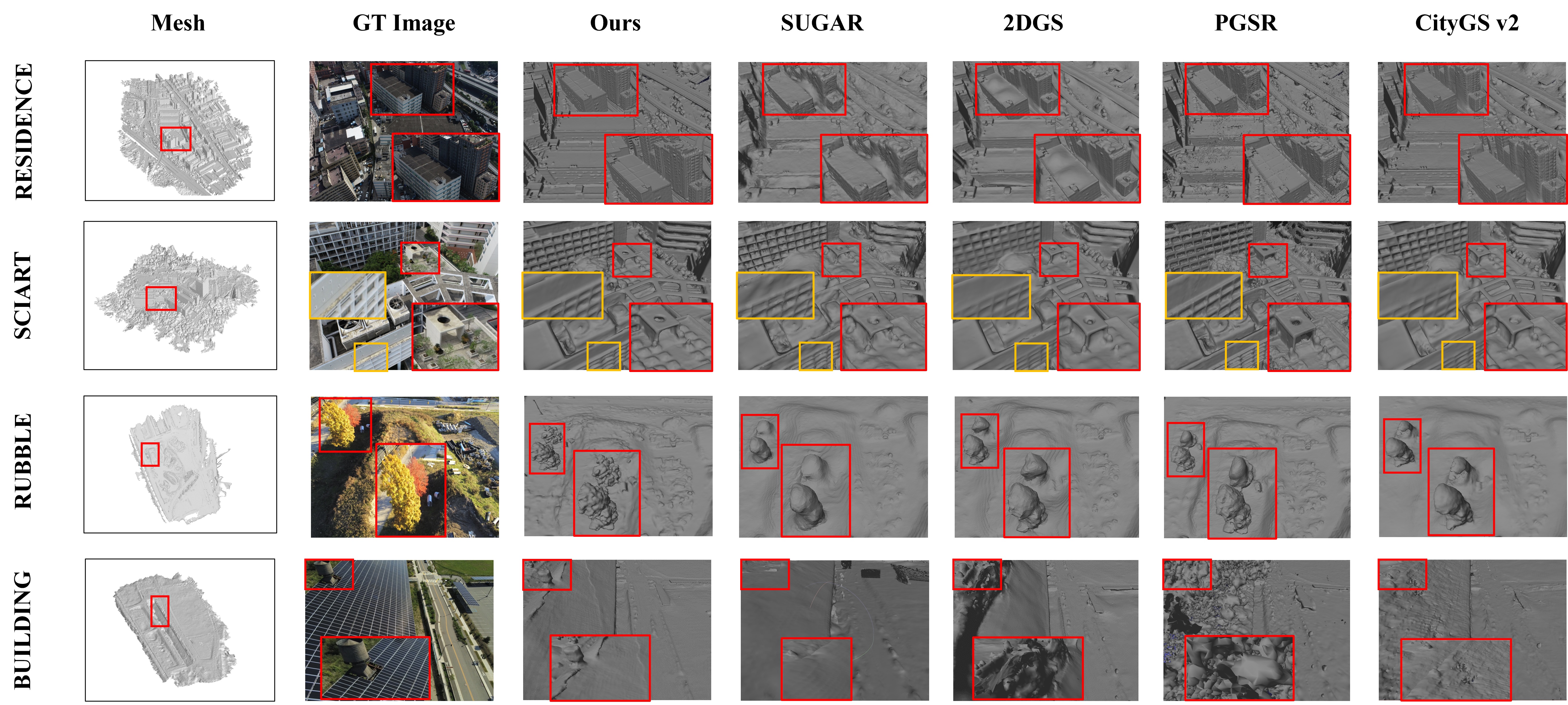}
    \caption{Local surface reconstruction details of different methods on  UrbanScene3D \citep{lin2022urbanscene3d} and Mill19 \citep{turki2022meganerf}. Representative regions are shown to compare geometric recovery in regular building structures, weakly textured surfaces, and complex vegetation areas.}
    \label{fig:urbanscene3d_mill19_local_details}
\end{figure*}

\paragraph{Results on UrbanScene3D and Mill19} Fig.~\ref{fig:urbanscene3d_mill19_complete_mesh} shows the full-scene reconstruction results on  UrbanScene3D \citep{lin2022urbanscene3d} and Mill19 \citep{turki2022meganerf}. In the UrbanScene3D scenes dominated by dense buildings, STARS-GS recovers the main structures of building groups with relatively complete geometry while preserving major geometric boundaries. In the Mill19 scenes containing more bare ground and vegetation, the reconstructed surfaces preserve the overall terrain variations as well as the irregular surface geometry of vegetation regions.

\begin{figure*}[t]
    \centering
    \includegraphics[width=\textwidth]{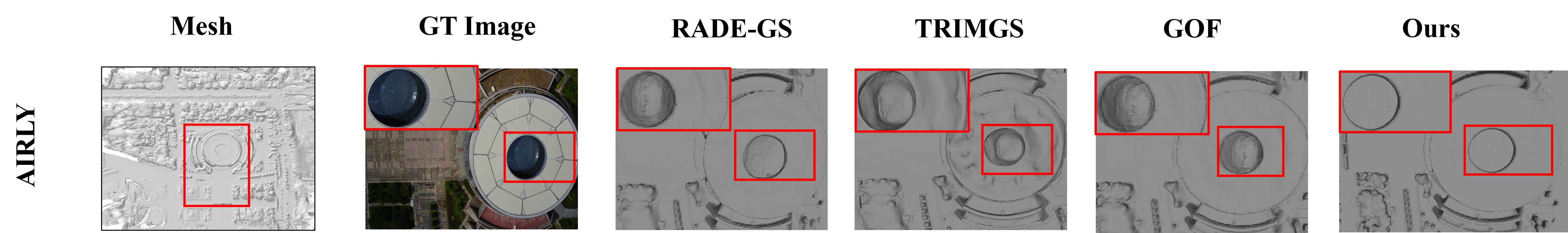}
    \caption{Local surface reconstruction results of STARS-GS and additional Gaussian surface reconstruction methods on AIRLY \citep{yao2026arsgaussian}. Representative local regions are shown to compare surface continuity, boundary completeness, and local geometric details.}
    \label{fig:extended_surface_comparison}
\end{figure*}

Fig.~\ref{fig:urbanscene3d_mill19_local_details} further compares local reconstruction results in representative regions of UrbanScene3D and Mill19 to examine whether the qualitative advantages observed in the primary evaluation persist across these additional datasets. In the Residence scene, the compared methods exhibit different local artifacts around roof structures and adjacent building facades, including over-smoothing, geometric thickening, local protrusions, floater artifacts, and blurred details. By contrast, STARS-GS more accurately recovers roof boundaries and facade structures with clearer local geometry. In the Sci-Art scene, STARS-GS more completely preserves slender columns and regularly arranged window grids, while maintaining the continuity of these fine structures. 

In Mill19, differences among the methods are more pronounced in the vegetation regions of the Rubble scene. STARS-GS avoids the over-smoothing observed in the compared methods and more accurately recovers the internal layered undulations and irregular boundaries of tree canopies. In the Building scene, STARS-GS also more clearly reconstructs locally protruding components on relatively planar surfaces, preserving their geometric distinction from the surrounding surface. Together, the results on UrbanScene3D and Mill19 show that STARS-GS preserves both regular building structures and the irregular geometry of vegetation surfaces, consistent with the results on GauU-Scene and AIRLY and indicating robust and consistent reconstruction across datasets with different scene characteristics.

\paragraph{Extended Comparison with Surface Reconstruction Methods} Some Gaussian-based surface reconstruction methods are not readily scalable to the large scenes considered in the main evaluation. Given its comparatively smaller scene area, AIRLY provides a more compact evaluation setting that allows direct comparison with these methods. We therefore conduct an extended comparison with GSDF, RaDe-GS, and Trim2DGS on AIRLY to further assess the surface reconstruction performance of STARS-GS against additional Gaussian-based methods.

Table~\ref{tab:extended_surface_comparison} reports the quantitative results on AIRLY, where STARS-GS again achieves the best Precision, Recall, and F1-score among all compared methods. Fig.~\ref{fig:extended_surface_comparison} further presents the reconstructed surfaces in representative local regions. GSDF, RaDe-GS, and Trim2DGS exhibit varying degrees of surface artifacts, boundary distortions, and loss of geometric details in these regions. By comparison, STARS-GS produces more continuous and complete surfaces with clearer structural boundaries.

\begin{table}[t]
\centering
\caption{Additional quantitative comparison with representative Gaussian-based surface reconstruction methods on AIRLY \citep{yao2026arsgaussian}. We report Precision, Recall, and F1-score, with the best and second-best results shown in \textbf{bold} and \underline{underlined}, respectively.}
\label{tab:extended_surface_comparison}
\small
\setlength{\tabcolsep}{4pt}
\renewcommand{\arraystretch}{1.10}
\begin{tabular*}{\linewidth}{@{\extracolsep{\fill}}lccc@{}}
\toprule
Method & Precision $\uparrow$ & Recall $\uparrow$ & F1-score $\uparrow$ \\
\midrule
GSDF
& 0.647 & 0.621 & 0.634 \\
RaDe-GS
& \underline{0.695}
& \underline{0.646}
& \underline{0.670} \\
Trim2DGS
& 0.672 & 0.615 & 0.642 \\
STARS-GS (Ours)
& \textbf{0.723}
& \textbf{0.716}
& \textbf{0.719} \\
\bottomrule
\end{tabular*}
\end{table}

\subsubsection{Novel View Synthesis}
\label{subsubsec:novel_view_synthesis}

\begin{figure*}[!t]
    \centering
    \includegraphics[width=\textwidth]{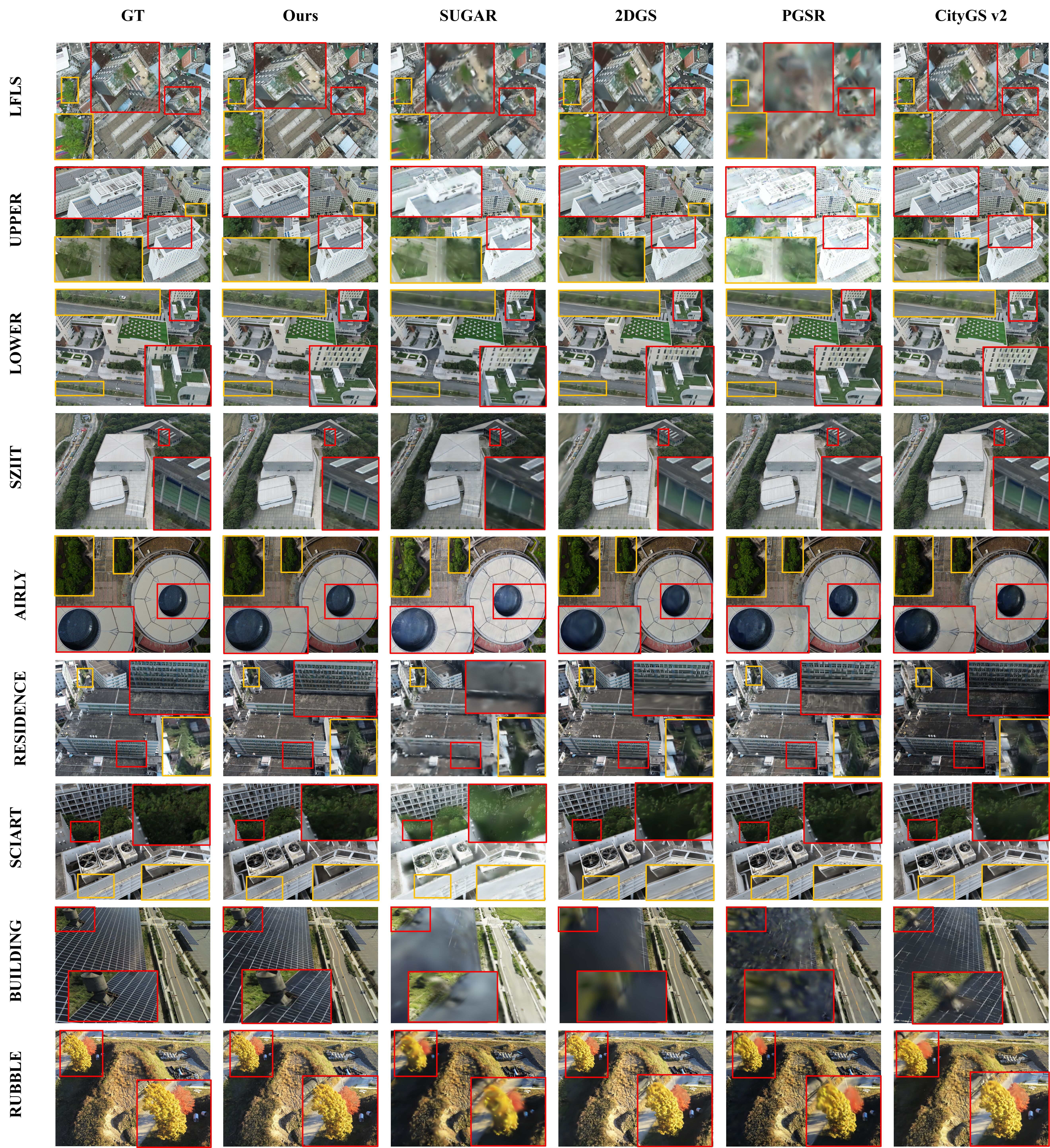}
    \caption{Qualitative novel view synthesis results of different methods in representative test views. The comparisons focus on building boundaries, roof textures, road boundaries, vegetation, and other high-frequency details.}
    \label{fig:nvs_qualitative}
\end{figure*}

We further compare the novel view synthesis performance of STARS-GS with the selected baseline methods to assess whether the geometric improvements are achieved without sacrificing rendering quality.

\paragraph{Main Comparison}

\begin{table*}[t]
\centering
\caption{Quantitative novel view synthesis comparison with the methods included in the main comparison on all test scenes from GauU-Scene \citep{xiong2024gauuscene}, UrbanScene3D \citep{lin2022urbanscene3d}, Mill19 \citep{turki2022meganerf}, and AIRLY \citep{yao2026arsgaussian}. We report PSNR, SSIM, and LPIPS, with the best and second-best results shown in \textbf{bold} and \underline{underlined}, respectively.}
\label{tab:nvs_main}

\footnotesize
\setlength{\tabcolsep}{1.5pt}
\renewcommand{\arraystretch}{1.08}

\begin{tabular*}{\textwidth}{@{\extracolsep{\fill}}cl*{9}{c}@{}}
\toprule
\multirow{2}{*}{Method}
& \multirow{2}{*}{Metric}
& \multicolumn{9}{c}{Scene} \\
\cmidrule(lr){3-11}
& & Lower & Upper & LFLS & SZIIT & Sci-Art & Residence
& Building & Rubble & AIRLY \\
\midrule

\multirow{3}{*}{SuGaR}
& PSNR  & 22.85 & 22.34 & 21.76 & 22.01 & 20.52 & 20.78 & 20.98 & 20.75 & 23.72 \\
& SSIM  & 0.703 & 0.694 & 0.681 & 0.688 & 0.624 & 0.636 & 0.655 & 0.638 & 0.721 \\
& LPIPS & 0.372 & 0.386 & 0.412 & 0.398 & 0.438 & 0.421 & 0.431 & 0.449 & 0.307 \\
\midrule

\multirow{3}{*}{2DGS}
& PSNR  & 24.36 & 24.08 & 23.41 & 23.72 & 22.18 & 22.36 & 22.74 & 22.42 & 24.78 \\
& SSIM  & 0.782 & 0.773 & 0.751 & 0.762 & 0.691 & 0.702 & 0.718 & 0.701 & 0.779 \\
& LPIPS & 0.296 & 0.305 & 0.332 & 0.318 & 0.354 & 0.338 & 0.353 & 0.367 & 0.247 \\
\midrule

\multirow{3}{*}{PGSR}
& PSNR
& \underline{25.18}
& \underline{25.16}
& \underline{24.36}
& 24.52
& 22.96
& \underline{23.21}
& \underline{23.56}
& 23.21
& \underline{25.62} \\
& SSIM
& \underline{0.817}
& 0.809
& \underline{0.792}
& 0.801
& 0.722
& \underline{0.735}
& \underline{0.755}
& 0.739
& \underline{0.816} \\
& LPIPS
& \underline{0.246}
& 0.258
& \underline{0.282}
& 0.269
& 0.292
& \underline{0.278}
& \underline{0.297}
& 0.314
& \underline{0.206} \\
\midrule

\multirow{3}{*}{CityGaussianV2}
& PSNR
& 25.07
& \underline{25.16}
& 24.18
& \underline{24.69}
& \underline{23.14}
& 23.08
& 23.48
& \underline{23.35}
& 25.48 \\
& SSIM
& 0.811
& \underline{0.816}
& 0.785
& \underline{0.807}
& \underline{0.731}
& 0.729
& 0.748
& \underline{0.744}
& 0.811 \\
& LPIPS
& 0.253
& \underline{0.249}
& 0.289
& \underline{0.261}
& \underline{0.286}
& 0.284
& 0.302
& \underline{0.306}
& 0.212 \\
\midrule

\multirow{3}{*}{STARS-GS}
& PSNR
& \textbf{25.71}
& \textbf{25.64}
& \textbf{24.92}
& \textbf{25.21}
& \textbf{23.82}
& \textbf{23.67}
& \textbf{24.15}
& \textbf{23.82}
& \textbf{26.21} \\
& SSIM
& \textbf{0.836}
& \textbf{0.831}
& \textbf{0.814}
& \textbf{0.824}
& \textbf{0.758}
& \textbf{0.764}
& \textbf{0.781}
& \textbf{0.762}
& \textbf{0.841} \\
& LPIPS
& \textbf{0.204}
& \textbf{0.211}
& \textbf{0.236}
& \textbf{0.224}
& \textbf{0.238}
& \textbf{0.229}
& \textbf{0.248}
& \textbf{0.259}
& \textbf{0.174} \\

\bottomrule
\end{tabular*}
\end{table*}

Table~\ref{tab:nvs_main} reports the novel view synthesis results of STARS-GS and the methods included in the main comparison under the unified evaluation protocol. Across all evaluated scenes, STARS-GS achieves the best PSNR, SSIM, and LPIPS. Its average PSNR and SSIM reach 24.79~dB and 0.801, respectively, outperforming the best-performing comparison method by 0.60~dB and 0.025. Its average LPIPS reaches 0.225, representing a relative reduction of 17.2\%. These results show that the improved geometric reconstruction quality of STARS-GS does not come at the expense of appearance quality.

Fig.~\ref{fig:nvs_qualitative} presents novel view synthesis results from representative test views. The differences are most apparent in high-frequency regions, including building boundaries, roof textures, road boundaries, and vegetation regions. Several comparison methods exhibit noticeable texture blurring, boundary smearing, and local appearance artifacts in these regions. In contrast, STARS-GS produces more realistic renderings, with clearer structural boundaries and better preservation of texture details.

\begin{figure}[!htbp]
    \centering
    \includegraphics[width=\linewidth]{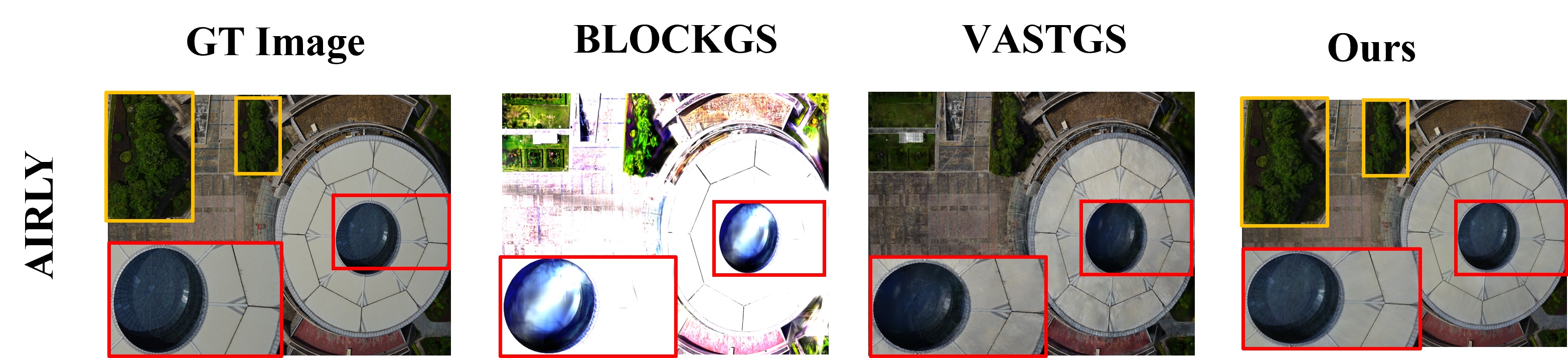}
    \caption{Supplementary qualitative novel view synthesis comparison with large-scene 3DGS methods on AIRLY \citep{yao2026arsgaussian}. Representative local regions are selected to compare color consistency, structural boundaries, and texture details.}
    \label{fig:nvs_additional}
\end{figure}

\paragraph{Supplementary Comparison with Large-Scene NVS Methods} Table~\ref{tab:nvs_additional} shows that STARS-GS achieves the best PSNR, SSIM, and LPIPS among the additional large-scene methods. Fig.~\ref{fig:nvs_additional} presents the corresponding rendering results. BlockGaussian exhibits noticeable exposure bias and color inconsistency, whereas VastGaussian produces texture smoothing and missing details around roads, vegetation, and building boundaries. By comparison, STARS-GS maintains more consistent appearance and preserves finer texture details.

\begin{table}[!htbp]
\centering
\caption{Supplementary quantitative novel view synthesis comparison with large-scene 3DGS methods on AIRLY \citep{yao2026arsgaussian}. We report PSNR, SSIM, and LPIPS, with the best and second-best results shown in \textbf{bold} and \underline{underlined}, respectively.}
\label{tab:nvs_additional}
\small
\setlength{\tabcolsep}{4pt}
\renewcommand{\arraystretch}{1.08}
\begin{tabular*}{\linewidth}{@{\extracolsep{\fill}}lccc@{}}
\toprule
Method & PSNR $\uparrow$ & SSIM $\uparrow$ & LPIPS $\downarrow$ \\
\midrule
BlockGaussian & 22.370 & 0.735 & 0.310 \\
VastGaussian & \underline{25.790} & \underline{0.829} & \underline{0.196} \\
STARS-GS (Ours) & \textbf{26.210} & \textbf{0.841} & \textbf{0.174} \\
\bottomrule
\end{tabular*}
\end{table}

\subsection{Ablation Study}
\label{subsec:ablation_study}

To examine the contribution of the key designs in STARS-GS, we use the full model as the reference and construct controlled variants by replacing or removing individual components at a time. Specifically, structure-aware partitioning is replaced with the camera-visibility-based partitioning of VastGaussian (Vast-GS)\citep{lin2024vastgaussian} and point-cloud-density-based partitioning of BlockGaussian (Block-GS)\citep{wu2025blockgaussian} and the effect of boundary refinement is evaluated by removing the overlap-band constraint. Neighborhood-aware organization is replaced by no explicit geometric constraint or conventional depth-and-normal constraints, and adaptive surface regularization is replaced by either no or uniformly weighted regularization. All variants use the same data splits, training configurations, and mesh-extraction protocol.

\begin{table*}[t]
\centering
\caption{ Quantitative ablation results on five scenes. Precision (P), Recall (R), and F1-score are reported; indented +'' rows denote alternative strategies after removing the corresponding design. }
\label{tab:ablation_main}
\footnotesize
\setlength{\tabcolsep}{1.2pt}
\renewcommand{\arraystretch}{1.10}
\begin{tabular*}{\textwidth}
{@{\extracolsep{\fill}}l*{15}{c}@{}}
\toprule
\multirow{2}{*}{Ablation Setting}
& \multicolumn{3}{c}{Lower}
& \multicolumn{3}{c}{Upper}
& \multicolumn{3}{c}{LFLS}
& \multicolumn{3}{c}{SZIIT}
& \multicolumn{3}{c}{AIRLY} \\
\cmidrule(lr){2-4}
\cmidrule(lr){5-7}
\cmidrule(lr){8-10}
\cmidrule(lr){11-13}
\cmidrule(lr){14-16}
& P $\uparrow$ & R $\uparrow$ & F1 $\uparrow$
& P $\uparrow$ & R $\uparrow$ & F1 $\uparrow$
& P $\uparrow$ & R $\uparrow$ & F1 $\uparrow$
& P $\uparrow$ & R $\uparrow$ & F1 $\uparrow$
& P $\uparrow$ & R $\uparrow$ & F1 $\uparrow$ \\
\midrule
w/o Structure-Aware Partitioning
& \multicolumn{15}{c}{} \\
\quad + VastGaussian
& 0.689 & 0.669 & 0.679
& 0.672 & 0.682 & 0.677
& 0.651 & 0.642 & 0.646
& 0.636 & 0.647 & 0.641
& 0.699 & 0.689 & 0.694 \\
\quad + BlockGaussian
& 0.680 & 0.654 & 0.667
& 0.663 & 0.666 & 0.664
& 0.642 & 0.625 & 0.633
& 0.627 & 0.631 & 0.629
& 0.688 & 0.671 & 0.679 \\
w/o Boundary Refinement
& 0.701 & 0.685 & 0.693
& 0.688 & 0.700 & 0.694
& 0.677 & 0.674 & 0.675
& 0.662 & 0.673 & 0.667
& 0.715 & 0.706 & 0.710 \\
\midrule
w/o Neighborhood-Aware Organization
& 0.646 & 0.625 & 0.635
& 0.640 & 0.651 & 0.645
& 0.611 & 0.600 & 0.605
& 0.593 & 0.604 & 0.598
& 0.659 & 0.642 & 0.650 \\
\quad + Depth \& Normal
& 0.665 & 0.649 & 0.657
& 0.658 & 0.674 & 0.666
& 0.632 & 0.628 & 0.630
& 0.617 & 0.629 & 0.623
& 0.681 & 0.670 & 0.675 \\
\midrule
w/o Adaptive Surface Regularization
& 0.675 & 0.655 & 0.665
& 0.662 & 0.682 & 0.672
& 0.646 & 0.638 & 0.642
& 0.629 & 0.642 & 0.635
& 0.690 & 0.676 & 0.683 \\
\quad + Uniform Regularization
& 0.700 & 0.676 & 0.688
& 0.685 & 0.691 & 0.688
& 0.666 & 0.652 & 0.659
& 0.648 & 0.658 & 0.653
& 0.706 & 0.683 & 0.694 \\
\midrule
\textbf{Full STARS-GS}
& \textbf{0.708} & \textbf{0.694} & \textbf{0.701}
& \textbf{0.695} & \textbf{0.712} & \textbf{0.703}
& \textbf{0.686} & \textbf{0.690} & \textbf{0.688}
& \textbf{0.673} & \textbf{0.689} & \textbf{0.681}
& \textbf{0.723} & \textbf{0.716} & \textbf{0.719} \\
\bottomrule
\end{tabular*}
\end{table*}

Table~\ref{tab:ablation_main} reports that the full STARS-GS achieves the best overall performance across all five scenes, whereas removing or replacing any component leads to varying degrees of performance degradation. This indicates that all proposed components contribute to the accuracy and completeness of surface reconstruction.

\subsubsection{Impact of Partitioning and Boundary Refinement}
\label{subsubsec:ablation_partition_boundary}

\begin{figure}[!ht]
    \centering
    \includegraphics[width=\linewidth]{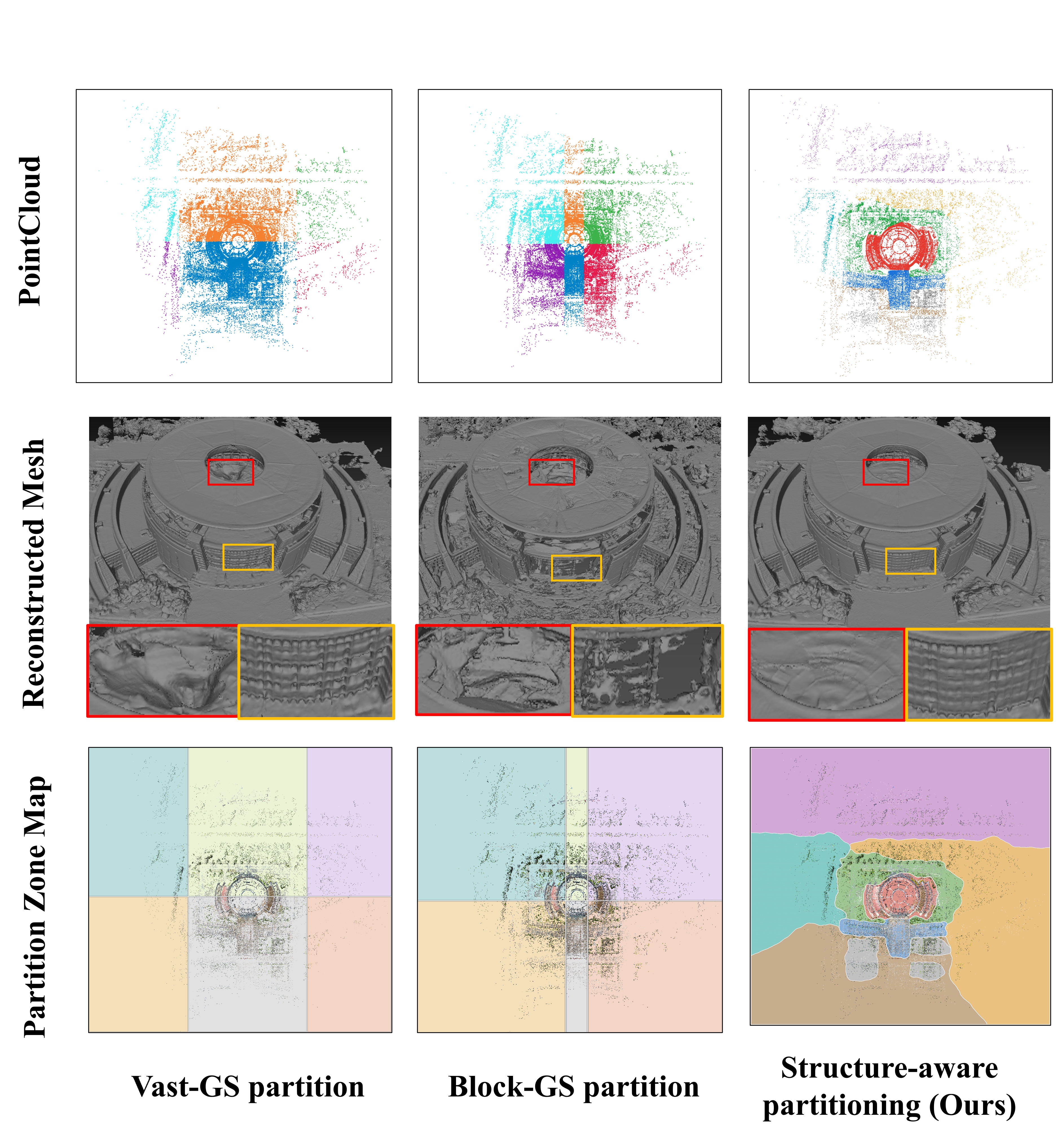}
    \caption{Ablation comparison of scene partitioning strategies. From left to right: Vast-GS, Block-GS, and STARS-GS, with corresponding meshes and enlarged local details.}
    \label{fig:partition_ablation}
\end{figure}

Replacing structure-aware scene partitioning decreases all three metrics across the five scenes, with larger drops in Recall, indicating its main contribution to surface completeness. As shown in Fig.~\ref{fig:partition_ablation} partitioning results, the horizontal or vertical boundaries produced by Vast-GS and Block-GS cut through the central building and split continuous structures across sub-regions, forcing the same scene element to be optimized independently, producing local depressions and surface disturbances on roofs and facades. In contrast, our structure-aware partitioning aligns boundaries with scene elements, preserving continuous structures and reducing roof undulations while recovering facade window grids more completely.

\begin{figure}[!ht]
    \centering
    \includegraphics[width=0.9\linewidth]{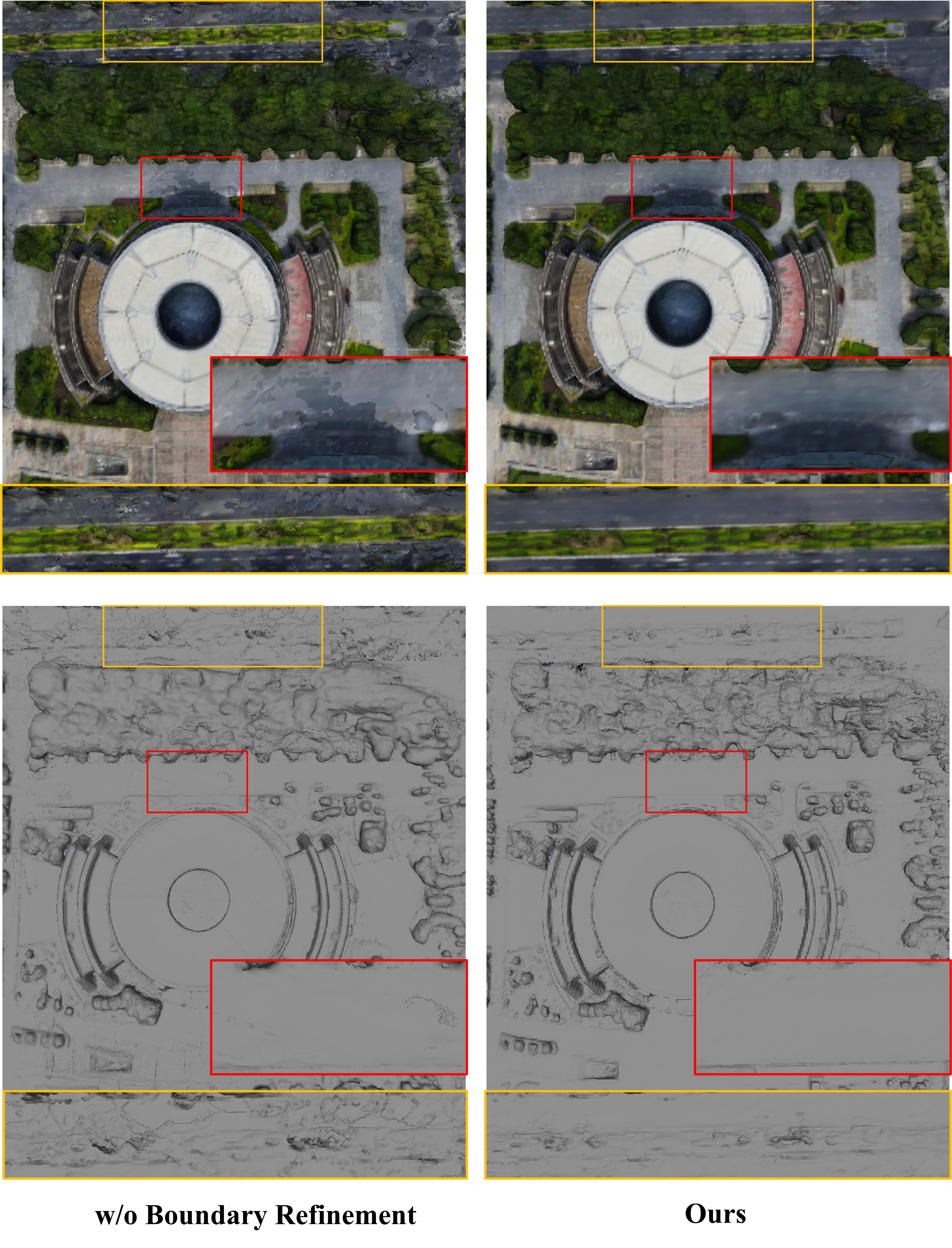}
    \caption{Ablation comparison of boundary refinement, with enlarged boundary regions showing stitching artifacts and cross-region geometric continuity.}
    \label{fig:overlap_ablation}
\end{figure}

Removing boundary refinement also reduces Precision and Recall, again with larger Recall drops. Fig.~\ref{fig:overlap_ablation} shows that, without this refinement, the textured mesh exhibits patch-like stitching artifacts, while the untextured mesh exhibits pronounced local geometric perturbations. The full model suppresses these artifacts and produces more continuous and complete surface geometry across adjacent sub-regions.

\subsubsection{Impact of Neighborhood-Aware Organization}
\label{subsubsec:ablation_topology}

Replacing neighborhood-aware Gaussian organization with either no explicit geometric constraints or depth and normal constraints causes the largest degradation among all ablations: the average F1-score drops from 0.698 to 0.627 without explicit geometric constraints and recovers only to 0.650 with conventional depth and normal constraints, but remains clearly below the full model. This indicates that the improvement comes not only from additional geometric supervision, but from neighborhood-aware primitive organization, which benefits both reconstruction accuracy and completeness.

\begin{figure}[!htbp]
    \centering
    \includegraphics[width=\linewidth]{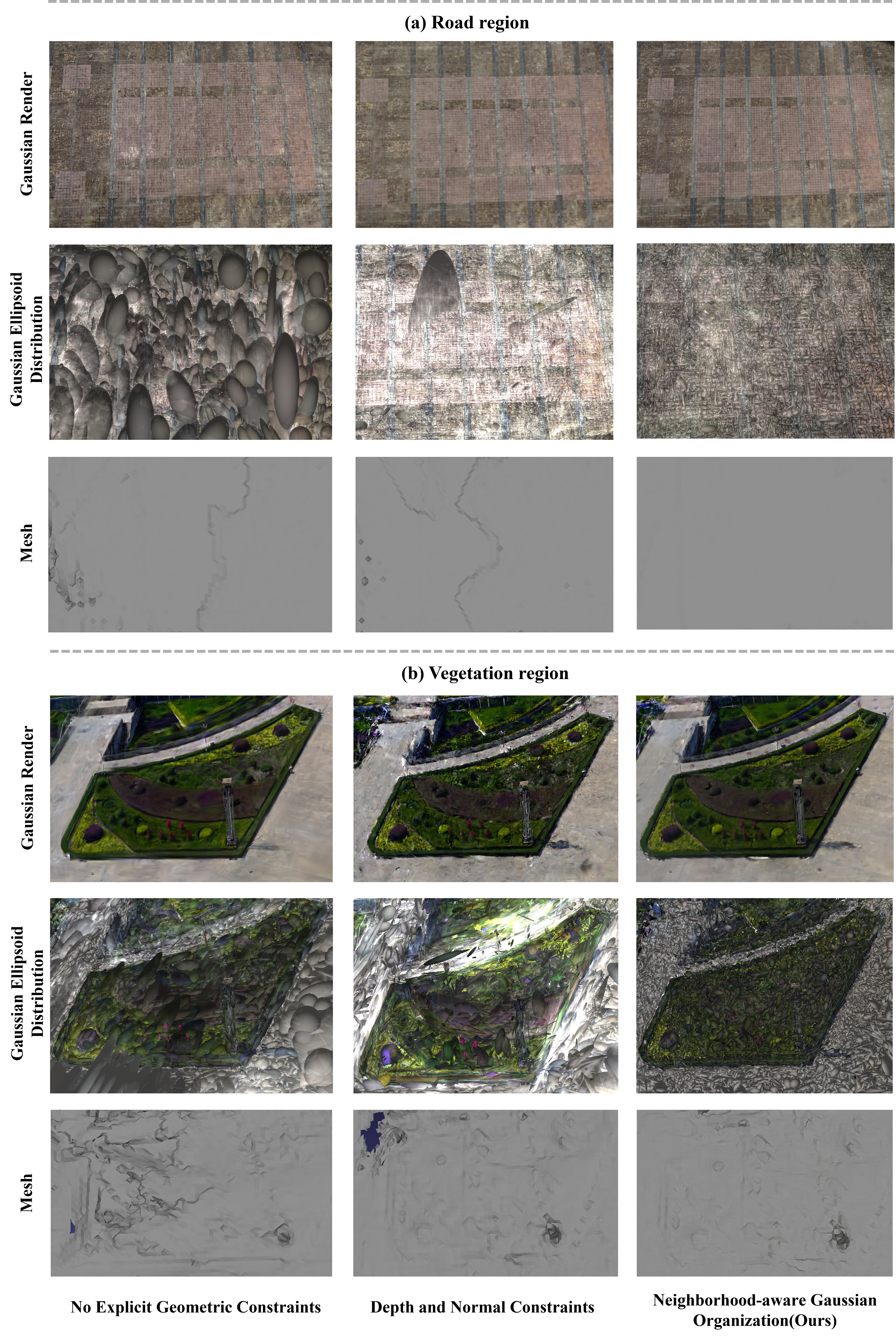}
    \caption{Ablation comparison of neighborhood-aware Gaussian organization on AIRLY, showing rendered images, Gaussian distributions, and meshes for different geometric constraints.}
    \label{fig:topology_ablation_gaussians}
\end{figure}

Fig.~\ref{fig:topology_ablation_gaussians} further compares Gaussian distributions and reconstructed geometry in representative road and vegetation regions. Although the three variants produce visually similar renderings, the Gaussian distributions and reconstructed meshes differ substantially, showing that comparable appearance quality does not necessarily imply well-organized geometry. In the road region, without explicit geometric constraints, many Gaussians become oversized and heavily overlap. Introducing depth and normal constraints makes most Gaussians flatter, but oversized and overlapping primitives remain and locally dominate the surface representation, resulting in irregular Gaussian organization and, consequently, mesh distortions. In contrast, neighborhood-aware organization encourages neighboring Gaussians to represent the surface jointly yielding a more balanced distribution that better conforms to the road geometry and, consequently, a flatter and more continuous mesh surface. A similar effect is observed in the vegetation region, where the more complex geometry amplifies primitive disorganization and subsequently leads to distortions or local holes. Neighborhood-aware organization better follows these surface variations while improving geometric completeness.

\begin{figure*}[t]
    \centering
    \includegraphics[width=\textwidth]{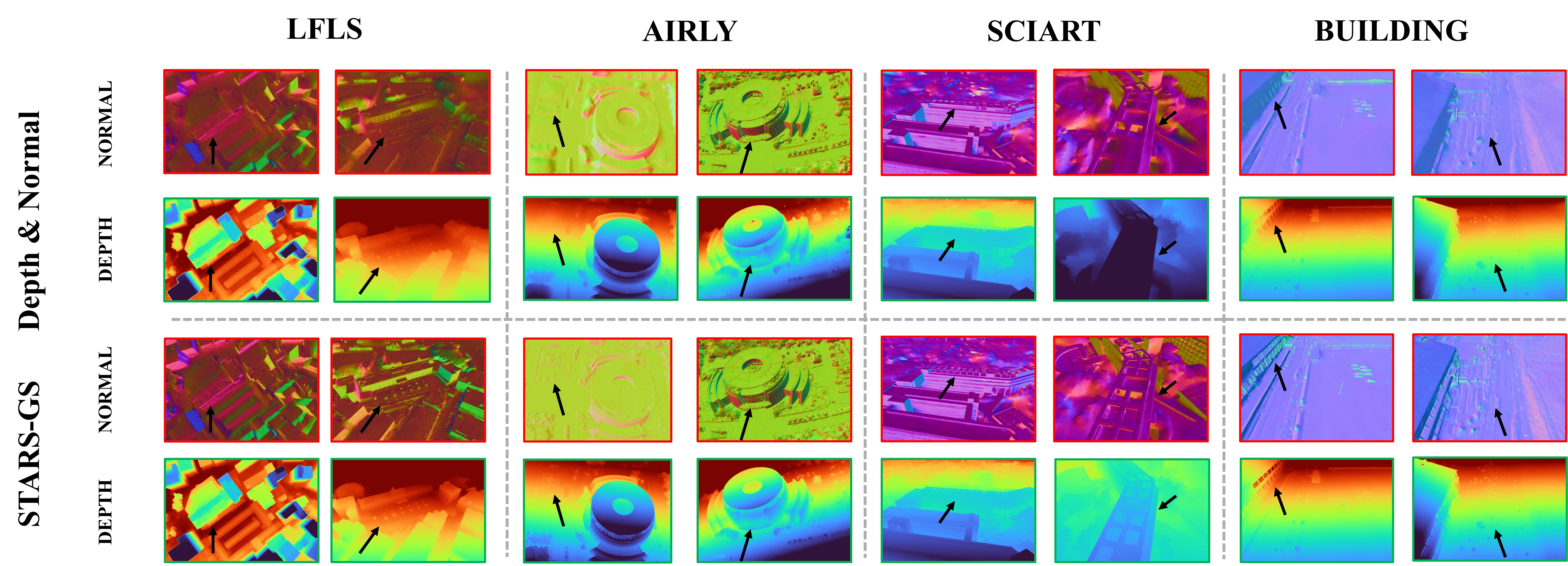}
    \caption{Cross-scene comparison of the depth-and-normal variant and full STARS-GS using normal and depth maps on LFLS, AIRLY, Sci-Art, and Building. Black arrows highlight differences in structural boundaries, local surface orientation, and depth separation}
    \label{fig:topology_ablation_geometry}
\end{figure*}

To verify whether these differences generalize across scenes, Fig.~\ref{fig:topology_ablation_geometry} compares the depth-and-normal variant with the full model using normal and depth maps. In LFLS of GauU-Scene, the depth-and-normal variant blurs building contours and depth separation between adjacent roofs and oversmooths orientation variations in vegetation. Similar degradation occurs on the circular architectural structures of AIRLY. The full model preserves clearer boundaries, clearer depth separation, and richer normal variations. In Sci-Art of UrbanScene3D, the depth-and-normal variant causes slender rooftop columns to merge with surrounding roof surfaces, with some columns distorted or partially missing. In Building of Mill19, the internal holes of hollow tubular supports are filled, and small lamp posts are also barely visible, as their geometry is excessively smoothed into the surrounding surface. Neighborhood-aware organization reduces such cross-surface interference, better separating slender structures from surrounding surfaces and preserving their geometry.

\subsubsection{Impact of Adaptive Surface Regularization}
\label{subsubsec:ablation_surface_regularization}

As shown in Table~\ref{tab:ablation_main}, removing surface regularization decreases the average F1-score from 0.698 to 0.659. Applying uniformly weighted surface regularization improves it to 0.676, but remains below the full model. This performance indicates that the benefit depends not only on surface regularization itself but also on adjusting its strength according to local geometric characteristics.

\begin{table}[t]
\centering
\caption{ Regional comparison of surface regularization on AIRLY. Regions are split by median planarity ($c_{\mathrm{med}}=0.47$), with P, R, and F1 reported. }
\label{tab:surface_regularization_region}
\small
\setlength{\tabcolsep}{2.0pt}
\renewcommand{\arraystretch}{1.10}
\begin{tabular*}{\linewidth}
{@{\extracolsep{\fill}}lcccccc@{}}
\toprule
\multirow{2}{*}{Regularization}
& \multicolumn{3}{c}{Low-planarity}
& \multicolumn{3}{c}{High-planarity} \\
\cmidrule(lr){2-4}
\cmidrule(lr){5-7}
& P $\uparrow$ & R $\uparrow$ & F1 $\uparrow$
& P $\uparrow$ & R $\uparrow$ & F1 $\uparrow$ \\
\midrule
None
& 0.649 & 0.692 & 0.670
& 0.731 & 0.646 & 0.686 \\
Uniform
& 0.612 & 0.584 & 0.598
& 0.749 & 0.688 & 0.717 \\
\textbf{Adaptive}
& \textbf{0.698}
& \textbf{0.779}
& \textbf{0.736}
& \textbf{0.770}
& \textbf{0.689}
& \textbf{0.727} \\
\bottomrule
\end{tabular*}
\end{table}

To examine this behavior across different local geometries, AIRLY is divided into low-planarity and high-planarity regions using the median planarity $c_{\mathrm{med}}=0.47$. Table~\ref{tab:surface_regularization_region} shows that uniform regularization affects the two region types differently. In high-planarity regions, uniform regularization improves F1 from 0.686 to 0.717, but remains below the adaptive result of 0.727. In low-planarity regions, however, it reduces F1 from 0.670 to 0.598, whereas adaptive regularization reaches 0.736. Thus, uniform regularization benefits high-planarity regions but can harm low-planarity geometry, whereas adaptive weighting remains effective in both, confirming its ability to balance structural consistency and local variation.

\begin{figure}[!htbp]
    \centering
    \includegraphics[width=\linewidth]{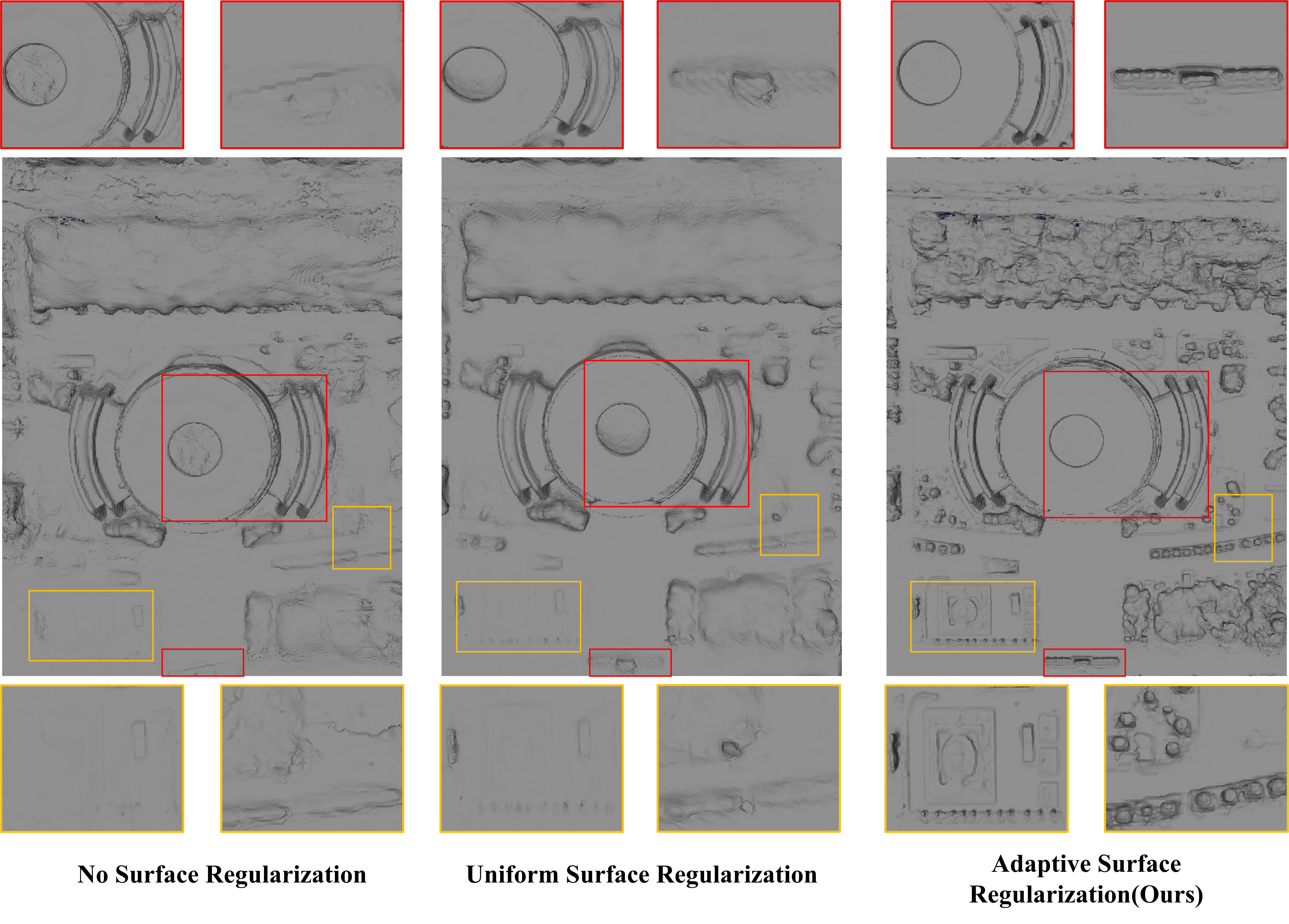}
    \caption{Comparison of surface regularization settings on AIRLY. The red and yellow boxes highlight high-planarity structured and low-planarity unstructured regions, respectively}
    \label{fig:surface_regularization_ablation}
\end{figure}

Fig.~\ref{fig:surface_regularization_ablation} compares the reconstruction results of three regularization settings on AIRLY, In the high-planarity structured region, removing surface regularization leads to surface irregularities and distorted boundaries around the circular roof. Uniform regularization suppresses some irregularities, but over-smooths local contours and small structures, making the roof boundary less clearly separated from adjacent structures. Adaptive surface regularization suppresses local surface irregularities while preserving clearer boundaries. In the low-planarity region, local structures are incomplete without regularization, whereas uniform regularization over-smooths valid geometric variations and blurs local features. By reducing the regularization strength, adaptive surface regularization better preserves these variations, producing clearer contours and more complete geometry.

\subsubsection{Parameter Sensitivity Analysis}
\label{subsubsec:parameter_sensitivity}

We evaluate the sensitivity of STARS-GS to the neighborhood size $k$, the maximum weight of the neighborhood organization loss $\lambda_{\mathrm{nbr}}^{\max}$, and the maximum weight of the adaptive surface regularization loss $\lambda_{\mathrm{surf}}^{\max}$. Each parameter is varied independently, with the others fixed. Results are reported in Table~\ref{tab:param_sensitivity}.

\begin{table}[t]
\centering
\caption{ Sensitivity analysis of key hyperparameters. Precision, Recall, and F1-score are reported; default settings are shown in \textbf{bold}.}
\label{tab:param_sensitivity}

\small
\setlength{\tabcolsep}{7pt}
\renewcommand{\arraystretch}{1.05}

\begin{tabular}{@{}ccccc@{}}
\toprule
Parameter & Value & Precision & Recall & F1-score \\
\midrule

\multirow{5}{*}{$k$}
& 10 & 0.684 & 0.675 & 0.679 \\
& 15 & 0.688 & 0.689 & 0.688 \\
& \textbf{20} & \textbf{0.697} & \textbf{0.700} & \textbf{0.698} \\
& 25 & 0.685 & 0.692 & 0.688 \\
& 30 & 0.676 & 0.688 & 0.682 \\

\midrule

\multirow{5}{*}{$\lambda_{\mathrm{nbr}}^{\max}$}
& 0.0025 & 0.676 & 0.687 & 0.681 \\
& 0.0050 & 0.685 & 0.692 & 0.688 \\
& \textbf{0.0100} & \textbf{0.697} & \textbf{0.700} & \textbf{0.698} \\
& 0.0200 & 0.686 & 0.689 & 0.687 \\
& 0.0400 & 0.674 & 0.680 & 0.677 \\

\midrule

\multirow{5}{*}{$\lambda_{\mathrm{surf}}^{\max}$}
& 0.0025 & 0.670 & 0.688 & 0.679 \\
& 0.0050 & 0.682 & 0.693 & 0.687 \\
& \textbf{0.0100} & \textbf{0.697} & \textbf{0.700} & \textbf{0.698} \\
& 0.0200 & 0.688 & 0.684 & 0.686 \\
& 0.0400 & 0.681 & 0.668 & 0.674 \\

\bottomrule
\end{tabular}
\end{table}

All three hyperparameters perform best near intermediate values. For the neighborhood size $k$, increasing it from 10 to 20 improves the F1-score from 0.679 to 0.698. Further increases reduce performance, suggesting that a small neighborhood provides insufficient local information, whereas an excessively large neighborhood may introduce weakly related connections and weaken local constraints. The two loss weights show similar trends: both $\lambda_{\mathrm{nbr}}^{\max}$ and $\lambda_{\mathrm{surf}}^{\max}$ achieve their best F1-score at 0.01. Smaller weights provide insufficient geometric regularization, whereas excessive constraints over-constrain local geometric variations.

Although overly small or overly large values lead to some performance degradation, the F1-score varies within relatively narrow ranges of 0.019, 0.021, and 0.024 for $k$, $\lambda_{\mathrm{nbr}}^{\max}$, and $\lambda_{\mathrm{surf}}^{\max}$, respectively, indicating that STARS-GS maintains relatively stable reconstruction performance across the evaluated parameter ranges.

\section{Conclusion}
\label{sec:conclusion}

In this work, we presented STARS-GS, a structure-aware 3DGS framework for large-scale surface reconstruction. Structure-aware scene partitioning organizes structurally coherent regions and refines shared boundaries to improve cross-region consistency. Neighborhood-aware Gaussian organization extends geometric constraints from individual primitives to their relative local organization, enabling neighboring Gaussians to better conform to local surface geometry. Adaptive surface regularization adjusts constraint strength to accommodate heterogeneous surface geometry.

Extensive experiments demonstrate improved geometric accuracy and surface completeness across large-scale scenes. STARS-GS achieves the highest F1-score on all quantitatively evaluated scenes while maintaining competitive training efficiency and model size. Qualitative comparisons further show that STARS-GS better preserves structural boundaries, local geometric details, and irregular surface variations across diverse scene types. STARS-GS also achieves the best PSNR, SSIM, and LPIPS, showing that the geometric improvements do not compromise appearance quality. Ablation studies further verify the effectiveness of each proposed component. 

Despite these improvements, several limitations remain. First, the current pipeline still relies on TSDF-based mesh extraction \citep{huang2024twodgs}. While this enables a unified geometric evaluation, the mesh extraction step itself may affect the final surface quality. Specifically, its finite resolution and multi-view averaging may attenuate fine geometric variations \citep{werner2014tsdf,li2022bnvfusion}. In addition, sparse or inconsistent depth observations may lead to incomplete surface geometry or inaccurate surface localization \citep{weder2020routedfusion,rosinol2023probabilistic}. Therefore, geometric errors in the final mesh do not necessarily originate entirely from the optimized Gaussian geometry. Future work could explore more direct surface extraction from optimized Gaussian representations to reduce geometry loss during mesh generation. Second, the performance of STARS-GS is partly determined by the reliability of the initial SfM reconstruction, including image coverage, camera poses, and sparse point quality \citep{wu2025sparse2dgs,fu2024colmapfree,xiang2026gaussiancraft}. Unreliable initialization may cause scene partitioning to deviate from the actual structure and the initialized neighborhood relationships to inaccurately reflect the local geometry. Future work could improve robustness through stronger geometric initialization, joint camera-pose optimization, and cross-view geometric constraints. Finally, while the proposed geometric designs improve reconstruction quality, they introduce additional computational overhead through geometric feature construction, neighborhood construction and repeated graph updates, as well as surface regularization. Although multi-GPU parallelization substantially reduces the wall-clock training time, it does not eliminate these computational costs. Future work could explore incremental graph updates, sparse neighborhood constraints, and more efficient parallel scheduling to improve computational efficiency while maintaining reconstruction quality.

\bibliographystyle{elsarticle-harv}
\bibliography{references}
\end{document}